\let\latexdocument\document
\let\latexenddocument\enddocument

\documentclass[shortpaper]{clv3}
\let\document\latexdocument
\let\enddocument\latexenddocument

\NewCommandCopy{\cnumdef}{\numdef}
\NewCommandCopy{\endcnumdef}{\endnumdef}
\let\numdef\relax \let\endnumdef\relax

\usepackage{amsmath}
\usepackage{amssymb}

\usepackage{tabularx}
\usepackage{multirow}
\usepackage{multicol}
\usepackage{dblfloatfix}
\usepackage{ragged2e} 
\usepackage{booktabs,makecell, multirow, tabularx}
\usepackage{subcaption}
\usepackage{booktabs}
\usepackage{float}
\usepackage{CJKutf8}

\usepackage{tikz}
\usetikzlibrary{calc,trees,positioning,arrows,chains,shapes.geometric,%
    decorations.pathreplacing,decorations.pathmorphing,shapes,%
    matrix,shapes.symbols}

\tikzset{
>=stealth',
  punktchain/.style={
    rectangle, 
    rounded corners, 
    draw=black, very thick,
    text width=30em, 
    minimum height=3em, 
    text centered, 
    on chain},
  line/.style={draw, thick, <-},
  element/.style={
    tape,
    top color=white,
    bottom color=blue!50!black!60!,
    minimum width=8em,
    draw=blue!40!black!90, very thick,
    text width=10em, 
    minimum height=3.5em, 
    text centered, 
    on chain},
  every join/.style={->, thick,shorten >=1pt},
  decoration={brace},
  tuborg/.style={decorate},
  tubnode/.style={midway, right=2pt},
}

\usepackage{hyperref}

\usepackage{xcolor}
\usepackage{color}
\definecolor{darkblue}{rgb}{0, 0, 0.5}
\hypersetup{colorlinks=true,citecolor=darkblue, linkcolor=darkblue, urlcolor=darkblue}

\begin{document}

\issue{1}{1}{2016}

\runningtitle{ArguGPT}

\runningauthor{Liu et.~al.}

\pageonefooter{Action editor: \{action editor name\}. Submission received: DD Month YYYY; revised version received: DD Month YYYY; accepted for publication: DD Month YYYY.}

\title{ArguGPT: evaluating, understanding and identifying argumentative essays generated by GPT models}

\author{Yikang Liu}
\affil{Shanghai Jiao Tong University}

\author{Ziyin Zhang}
\affil{Shanghai Jiao Tong University}

\author{Wanyang Zhang}
\affil{Peking University}

\author{Shisen Yue}
\affil{Shanghai Jiao Tong University}

\author{Xiaojing Zhao}
\affil{Shanghai Jiao Tong University}

\author{Xinyuan Cheng}
\affil{Shanghai Jiao Tong University}

\author{Yiwen Zhang}
\affil{Amazon}

\author{Hai Hu\thanks{corresponding author: hu.hai@sjtu.edu.cn}}
\affil{Shanghai Jiao Tong University}

\maketitle

\begin{abstract}
Content generated by artificial intelligence models have presented considerable challenge to educators around the world. 
When students submit AI generated content (AIGC) as their own work,  instructors will need to be able to detect such text, either with the naked eye or with the help of computational tools.  
There is also growing need and interest to understand the lexical, syntactic and stylistic features of AIGC among computational linguists. 

To address these challenges in the context of argumentative essay writing, we present ArguGPT, a carefully balanced corpus of 4,038 argumentative essays generated by 7 GPT models in response to essay prompts from three sources: (1) in-class or homework exercises, (2) TOEFL writing tasks and (3) GRE writing tasks. These machine-generated texts are paired with roughly equal number of human-written essays with low, medium and high scores matched in essay prompts. We also include an out-of-distribution test set where the machine essays are generated by models other than the GPT family---claude-instant, bloomz and flan-t5---to examine AIGC detectors' generalization ability. 

We then hire English instructors to distinguish machine essays from human ones. Results show that when first exposed to machine-generated essays, the instructors only have an accuracy of 61 percent in detecting them. 
But the number rises to 67 percent after one round of minimal self-training.
Next, we perform linguistic analyses of the machine and human essays, which
show that machines produce sentences with more complex syntactic structures while human essays tend to be lexically more complex.
Finally, we test existing AIGC detectors and build our own detectors using SVMs as well as the RoBERTa model. Our results suggest that a RoBERTa fine-tuned with the training set of ArguGPT can achieve above 90\% accuracy in 
\textcolor{black}{document, paragraph}
and sentence-level classification.  
\textcolor{black}{The document-level RoBERTa can generalize to other models as well (such as claude-instant), while off-the-shelf detectors such as GPTZero
fail to generalize to our out-of-distribution data.}

To the best of our knowledge, this is the first comprehensive analysis of argumentative essays produced by generative large language models.  Our work  demonstrates the need for educators to acquaint themselves with AIGC, presents the characteristics of AI generated argumentative essays and shows that detecting AIGC from the same domain seems to be an easy task for machine-learning based classifiers while transferring to essays generated by other models is challenging.
Machine-authored essays in ArguGPT and our models are publicly available at \url{https://github.com/huhailinguist/ArguGPT}. 
\end{abstract}

\section{Introduction}\label{sec:intro}

Recent large language models (LLM) such as ChatGPT have shown incredible generative abilities.
They have created many opportunities, as well as challenges for students and educators around the world.\footnote{This paper is written and polished by humans, with the exception of Appendix~\ref{app:instructions:human:exp}. 
} 
While students can use them to obtain information and increase efficiency in learning, many educators are concerned that ChatGPT will make it easier for students to cheat in their homework assignments, for example, by asking ChatGPT to summarize their readings, solve math problems, and even write responses and essays: tasks that are supposed to be completed by students themselves.
Educators have started to find that well-written essays submitted by students, some even deemed ``the best in class'', were actually written by ChatGPT, making it increasingly difficult to evaluate students' performance in class.
For instance, a philosophy professor from North Michigan University has discovered that the best essay in his class was in fact 
written by ChatGPT.\footnote{See \url{https://www.nytimes.com/2023/01/16/technology/chatgpt-artificial-intelligence-universities.html}} 
Thus it is critical for educators 
to identify AI generated content (AIGC), either with the naked eye, or the help of some tools.

\textcolor{black}{There is also growing interest and need among computational linguists to study texts generated by language models. Several studies have examined whether humans can identify AI-generated text~\citep{brown2020language-gpt3,clark_all_2021,dou_is_2022}. Others have built text classifiers to distinguish AI-written from human-written text~\citep{gehrmann2019gltr,mitchell2023detectgpt,guo2023close}.
}

The focus of this paper is on argumentative essays in the context of English as Other or Second Language (EOSL). There are an estimated 2 billion people learning/speaking English, and at least 12 million instructors worldwide\footnote{Data released by British Council in 2013 (See \url{https://www.britishcouncil.org/research-policy-insight/policy-reports/the-english-effect)}}. 
It is thus of practical significance for EOSL instructors to be able to identify AIGC; \textcolor{black}{ computational linguists also need to build efficient educational applications that accurately detect AI-generated essays. 
}

Therefore, our first goal is to establish a baseline of the performance of the EOSL instructors in distinguishing AIGC from texts written by non-native speakers.
We also want to examine whether their accuracy could be improved with minimal training. 
Next, we analyze the linguistic features of AIGC which contributes to a growing body of literature on AI-generated text~\citep{dou_is_2022,guo2023close}.
Last but not least, we aim to build and evaluate the performance of machine-learning classifiers on detecting AIGC.

Concretely, we ask the following questions:

\begin{itemize}
    \item Can human evaluators (language teachers) distinguish argumentative essays in English generated by GPT models from those written by human language learners?
    \item What are the linguistic features of machine-generated essays, compared with essays written by language learners?
    \item Can machine learning classifiers distinguish machine-generated essays from human-written ones? 
\end{itemize}

To answer these questions, we first collect 4,038 machine-generated essays using seven models of the GPT family (GPT2-XL, variants of GPT3, and ChatGPT), in response to 632 prompts from multiple levels of English proficiency and writing tasks (in-class writing exercises, TOEFL and GRE).
We then pair these essays with 4,115 human-written ones at low, medium and high level to form the ArguGPT corpus. 
\textcolor{black}{Also, an out-of-distribution test set is collected to evaluate the generalization ablity of our detectors, containing 500 machine essays and 500 human essays.}
We conduct human evaluation tests by asking 43 novice and experienced English instructors in China to identify whether a text is written by machine or human.
Next, we compare the 31 syntactic and lexical linguistic measures of human-authored and machine-generated essays, using the tools and methods from \citet{Lu2010AutomaticAO,Lu2012TheRO}, aiming to uncover the textual characteristics of GPT-generated essays.
Finally, we benchmark existing AIGC detectors such as GPTZero\footnote{\url{https://gptzero.me/}} and our own detectors based on SVM and RoBERTa on the development and test sets of the ArguGPT corpus.

Our major findings and contributions are:
    (1) We provide the first large-scale,  balanced corpus of AI-generated argumentative essays for NLP and ESOL researchers. 
    (2) We show that it is difficult for English instructors to identify GPT-generated texts. English instructors distinguish human- and GPT-authored essays with an accuracy of 61.6\% in the first round, and after some minimal training, the accuracy rises to 67.7\%, roughly 10 points higher than previously reported in \citet{clark_all_2021}, probably due to the instructors' familiarity with student-written texts.  Interestingly, they are better at detecting low-level human essays but high-level machine essays. 
    (3) In terms of syntactic and lexical complexity, we find that the best GPT models produce syntactically more complex sentences than humans (English language learners), but GPT-authored essays are often lexically less complex.
    (4) We discover that machine-learning classifiers can easily distinguish between machine-generated and human-authored essays, usually with very high accuracy, similar to results from \citet{guo2023close}. GPTZero has 90+\% accuracy both at the essay level and sentence level. Our best performing RoBERTa-large model finetuned on ArguGPT achieves 99\% accuracy on the test set at the essay level and 93\% at the sentence level.
    (5) \textcolor{black}{However, on the out-of-distribution test set, only the RoBERTa model finetuned on ArguGPT corpus shows consistent transfer learning ability. Performance of the two off-the-shelf detectors dropped dramatically, especially on detecting essays written by models not included in ArguGPT, e.g., gpt-4 and claude-instant. }
    (6) The machine-authored essays\footnote{As we do not have copyright to the human-authored essays, we will only release the index of the human essays use in our study. Interested readers can purchase relevant corpora from their owners to reproduce our results.  } will be released at \url{https://github.com/huhailinguist/ArguGPT}. Demo of our ArguGPT detector and related models are/will be available at \url{https://huggingface.co/spaces/SJTU-CL/argugpt-detector}.

This paper is structured in the following manner. 
Section \ref{sec:corpus} introduces the compilation process of the ArguGPT corpus. 
Section \ref{sec:human-eval} describes the method used for conducting human evaluation. 
Section \ref{sec:ling-ana} presents the linguistic analysis we conduct from syntactic and lexical perspectives. 
Section \ref{sec:classifier} discusses the performance of existing AIGC detectors and our own detectors on the test set of ArguGPT.
Section \ref{sec:related} introduces related works on Large Language Models (LLMs), human evaluation of AIGC, AIGC detectors, and AIGC's impact on education. 
Section \ref{sec:conclusion} concludes the paper.

\section{The ArguGPT corpus}

\label{sec:corpus}

In this section, we describe how we compile the ArguGPT corpus and an out-of-distribution (OOD) dataset for the evaluation of generalization ability. 

\subsection{Description of the ArguGPT corpus}

The ArguGPT corpus contains 4,115 human-written essays and 4,038
machine-generated essays produced by 7 GPT models. 
These essays are responses to prompts from three sources: (1) in-class writing exercises (WECCL, Written English Corpus of Chinese Learners), (2)  independent writing tasks in TOEFL exams (TOEFL11), and (3) the issue writing task in GRE (GRE) (see Table~\ref{tab:desc:corpus}). 
We first collect human essays from two existing corpora---WECCL~\citep{wen2005spoken} and TOEFL11~\citep{blanchard2013toefl11}, and compile a corpus for human-written GRE essays from GRE-prep materials ourselves. 
Next, we use the essay prompts from the three corpora of human essays to generate essays from seven GPT models listed in Table~\ref{tab:gen:info}.  
Example essay prompts can be found in Table~\ref{tab:ex:prompts}. 

\textcolor{black}{Here we list the characteristics of the ArguGPT corpus: 
\begin{itemize}
\setlength\itemsep{.1em}
    \item The human-written and machine-generated portions are comparable and matched in several respects: the number of the essays, the corpus size in tokens, the mean length of the essays, and the levels of the essays.
    \item Each machine-generated essay comes with a score given by an automated scoring system. (see Section \ref{sec:auto:score} for details). 
    \item The essays cover different levels of English proficiency. WECCL essays are written by low and intermediate English learners; TOEFL contains essays from all levels of writing ability; GRE essays are model essays written by highly proficient language users. 
    \item The human-written GRE essays are, to the best of our knowledge, the first corpus of GRE essays.
\end{itemize}
}

\begin{table}[t]
    \centering
    \resizebox{\textwidth}{!}{
    \begin{tabular}{cccccccc}\toprule
        Sub-corpus & \# essays & \# tokens  & mean len & \# prompts & \# low level & \# mid level & \# high level  \\\midrule
        WECCL-human & 1,845 & 450,657 & 244 & 25 & 369 & 1,107 & 369\\
        WECCL-machine & 1,813 & 442,531 & 244 & 25 & 281 & 785 & 747 \\\hline
        TOEFL11-human & 1,680 & 503,504 & 299 & 8 & 336 & 1,008 & 336\\
        TOEFL11-machine & 1,635 & 442,963 & 270 & 8 & 346 & 953 & 336 \\\hline
        GRE-human & 590 & 341,495 & 578 & 590 & 6 & 152 & 432\\
        GRE-machine & 590 & 268,640 & 455 & 590 & 2 & 145 & 443\\\hline
        OOD-human & 500 & 132,902 & 266 & - & - & - & - \\
        OOD-machine & 500 & 180,120 & 360 & - & - & - & - \\\midrule
        Total (w/o OOD) & 8,153 & 2,449,790 & 300 & 623 & 1,340 & 4,150 & 2,663\\
        \bottomrule
    \end{tabular}}
    \caption{Information of our ArguGPT corpus. All essays come with three levels of writing: $low$, $mid$, or $high$ $level$, scored using an automated system, with the exception of TOEFL11-human, where the scores are provided by owners of TOEFL11.}
    \label{tab:desc:corpus}
\end{table}

\begin{table}[t]
    \centering
    \resizebox{\textwidth}{!}{
    \begin{tabular}{c|p{0.85\textwidth}}
        \toprule
        Sub-corpus & Example Essay Prompt \\\midrule
        \multirow{4}{*}{WECCL} & Education is expensive, but the consequences of a failure to educate, especially in an increasingly globalized world, are even more expensive. \\\cline{2-2}
        \multirow{4}{*}{} & Some people think that education is a life-long process, while others don't agree. \\\hline
        \multirow{3}{*}{TOEFL11} & It is better to have broad knowledge of many academic subjects than to specialize in one specific subject. \\\cline{2-2}
        \multirow{3}{*}{} & Young people enjoy life more than older people do. \\\hline
        \multirow{4}{*}{GRE} & Major policy decisions should always be left to politicians and other government experts. \\\cline{2-2}
        \multirow{4}{*}{} & The surest indicator of a great nation is not the achievements of its rulers, artists, or scientists, but the general well-being of all its people. \\\bottomrule
    \end{tabular}}
    \caption{Example essay prompts. Note that we did not directly fed those to the models; see \ref{sec:prompt-sel} for more details about the \textit{full} prompts.}
    \label{tab:ex:prompts}
\end{table}

To the best of our knowledge, this is the first large-scale, prompt-balanced corpus of human and machine written English argumentative essays, with automated scores. We believe it can be beneficial to ESOL instructors, corpus linguists, and AIGC researchers.

\subsection{Collecting human-written essays}
\label{sec:collect:human}

Our goal is to collect human essays representing different levels of English proficiency. 
To this end, we decided to include essays from three sources: (1) WECCL, which we believe is representative of the writings from low to intermediate EOSL learners, (2) TOEFL11, representative of intermediate to advanced learners, and (3) GRE, which represents more advanced learners as well as native speakers. 
We elaborate on how the essays are collected and sampled below. 

\subsubsection{WECCL}

WECCL (Written English Corpus of Chinese Learners) corpus is the sub-corpus of SWECCL (Spoken and Written English Corpus of Chinese Learners)~\citep{wen2005spoken}. Texts in WECCL are essays written by English learners from Chinese universities, %
collected in the form of in-class writing tasks or after-class writing assignments. WECCL contains exposition essays and argumentative essays, but
only argumentative essays are used in our corpus. The original WECCL corpus has 4,678 essays in response to 26 prompts. We score
these essays with aforementioned automated scoring system, and then categorize them into three levels: low (score $\leq$ 13), medium (14 $\leq$ score $\leq$ 17), and high (score $\geq$ 18). 
To keep it balanced with the TOEFL subcorpus, we down-sample WECCL into 1,845 essays, ensuring the ratio of low:medium:high is 1:3:1.
From Table~\ref{tab:desc:corpus}, we can see that WECCL essays are shorted in length among the three human sub-corpora, with mean length 244 words per essay. 

\subsubsection{TOEFL11}

We use the TOEFL11 corpus released by ETS~\citep{blanchard2013toefl11}, which includes 12,100
essays written for the independent writing task in the Test of English as Foreign Language (TOEFL) 
by English learners with 11 native languages in response to 8 prompts. Since the essays come with 
three score levels (i.e., low, medium, high), we do not score them using the YouDao system. 
We down-sample TOEFL11 to 1,680 essays, making sure that we have the same number of essays per prompt. The ratio of low, medium, high is set to 1:3:1 as well.

\subsubsection{GRE} 

We also collect essays in response to the GRE \textit{issue} task. 
The Graduate Record Exam (GRE) has two writing tasks. The \textit{issue} task asks the test taker to write an essay on a specific issue whereas the \textit{argument} task requires the test take to read a text first and analyze the argument presented in the text mainly from logical aspect\footnote{More information about GRE writing could be found: \url{https://www.ets.org/gre/test-takers/general-test/prepare/content/analytical-writing.html}}. In keeping with the prompts of WECCL and TOEFL11, we only consider the \textit{issue} task in GRE.

As there are no publicly available corpus of GRE essays, we first collected 981 human written essays from 14 GRE-prep materials. An initial inspection shows that the collected essays have following two problems: 
1) some prompts do not conform to the usual GRE writing prompts (e.g., some prompts have only one phrase: ``Imaginative works vs. factual accounts''),
2) some essays show up in different GRE-prep materials.
After removing the problematic prompts and keeping only one of the reduplicated essays, a total of 590 essays remained. 
We then score these essays using the YouDao automatic scoring system, and assign essays to three levels: low (score $<$ 3), medium (3 $\leq$ score $<$ 5) and high (score $\geq$ 5).

Note that as these essays are sample essays from humans, only 6 out of the 590 essays are grouped into low level (see Table~\ref{tab:desc:corpus}). 

\subsection{Automatic scoring of essays}
\label{sec:auto:score}

We use automated scoring systems to score the essays in ArguGPT for two reasons: (1) to allow balanced sampling from different levels of human essays, and (2) to estimate the quality of machine essays generated by different models.\footnote{We did not score human TOEFL essays as they come with a three-level (low, medium and high) score from the TOEFL11 corpus~\citep{blanchard2013toefl11}.} 

In the pilot study, we use two automated scoring systems (YouDao and  Pigai\footnote{YouDao: \url{https://ai.youdao.com}; Pigai: \url{http://www.pigai.org/}}) to score a total of 480 machine essays on 10 prompts with 6 GPT models. Analyses show that the scores given by the two systems are highly correlated:  we see a Pearson correlation of 0.7570 for all 480 essays, 0.8730 when scores are grouped by prompts, and 0.9510 when grouped by models. 
Thus we decide to use only one system---YouDao, which provides an API for easy scoring. 
We further experiment with different settings of the YouDao system and decide to use their 30-point scale for TOEFL, 6-point scale for GRE, as they are optimized for TOEFL and GRE writing tasks, and a 20-point scale for WECCL, as our experiments show that this scale is most discriminating for essays generated by different models responding to WECCL prompts (see Appendix \ref{ap:auto:crit} for details).

\subsection{Collecting machine-generated essays}

In this section, we introduce how we collect machine-generated essays. 
We conduct minimal prompt-tuning to select a proper format of prompt according to scores given by the automated scoring system.
Finally, we use those prompts to generate essays.

\subsubsection{Prompt selection}
\label{sec:prompt-sel}

GPT models are prompt-sensitive~\citep{chen2023robust}. 
Thus for this study, we perform prompt tuning in our pilot.

We distinguish \textbf{essay prompt} from \textbf{added prompt}. An essay prompt is the sentence(s) that the test taker should respond to, e.g., ``Young people enjoy life more than older people do.''
An added prompt is the prompt or instruction added by us to prompt the model, e.g., ``Please write an essay of 400 words.''
One example is shown in Table \ref{tab:prompt:format}.

\begin{table}[H]
\centering
\resizebox{\textwidth}{!}{
\begin{tabular}{c|l}
\toprule
Prompt Type & Example \\\midrule
added prompt (\textit{prefix}) & Do you agree or disagree with the following statement? \\
essay prompt & Young people enjoy life more than older people do. \\
added prompt (\textit{suffix}) & Use specific reasons and examples to support your answer.\\\bottomrule
\end{tabular}}
\caption{In this example, we provide the added prompt that surrounds the essay prompt with both \textit{prefix} and \textit{suffix}. }
\label{tab:prompt:format}
\end{table}

An added prompt consists of two parts as the prefix and suffix to the essay prompt. 
Yet the prefix part is optional in our experimental settings.
Therefore, the full prompt given to the machines is in the following format, where sometimes only the suffix part of the added prompt is used: $$<Added\ prompt\ prefix> + <Essay\ prompt> + <Added\ prompt\ suffix>$$

Our goal is to find the best-added prompt that maximizes the scores given by the YouDao automated system. To this end, 
we first devise 20 added prompts and manually inspect the generated essays, which are then narrowed down to 5 added prompts that produce good essays. 
Next, we generate essays using each of the 5 added prompts and 2 essay prompts from each of the WECCL, TOEFL11 and GRE sub-corpus. 
The mean score of the essays generated by these prompts are shown in Table~\ref{tab:pmt:sel}.

\begin{table}[ht]
    \centering
    \resizebox{\textwidth}{!}{
    \begin{tabular}{cp{0.65\textwidth}ccc}
        \toprule
        No. & Content of the Added Prompt & TOEFL11 & WECCL & GRE\\
        \midrule
        \multirow{3}{*}{1} & Do you agree or disagree? Use specific reasons and examples to support your answer. Write an essay of roughly 300/400/500 words.  & \multirow{3}{*}{20.53} & \multirow{3}{*}{\textbf{20.56}} & \multirow{3}{*}{\textbf{20.97}} \\ \hline
        \multirow{2}{*}{2} & Do you agree or disagree? It is a test for English writing. Please write an essay of roughly 300/400/500 words. & \multirow{2}{*}{19.68} & \multirow{2}{*}{20.09} & \multirow{2}{*}{20.21} \\ \hline
        \multirow{4}{*}{3} & Do you agree or disagree? Pretend you are the best student in a writing class. Write an essay of roughly 300/400/500 words, with a large vocabulary and a wide range of sentence structures to impress your professor. & \multirow{4}{*}{20.41} & \multirow{4}{*}{19.71} & \multirow{4}{*}{19.65} \\ \hline
        \multirow{3}{*}{4} & Do you agree or disagree? Pretend you are a professional American writer. Write an essay of roughly 300/400/500 words, with the potential of winning a Nobel prize in literature. & \multirow{3}{*}{20.09} & \multirow{3}{*}{20.52} & \multirow{3}{*}{19.79} \\ \hline
        \multirow{3}{*}{5} & Do you agree or disagree? From an undergraduate student's perspective, write an essay of roughly 300/400/500 words to illustrate your idea. & \multirow{3}{*}{\textbf{20.65}} & \multirow{3}{*}{20.32} & \multirow{3}{*}{19.99} \\ 
        \bottomrule
    \end{tabular}}
    \caption{Mean scores for each added prompt in our pilot study.}
    \label{tab:pmt:sel}
\end{table}

From Table \ref{tab:pmt:sel}, we 
see that essays from different prompts seem to be very close on their scores. 
Thus we choose prompt 01 because it has the highest average score for the three subcorpora
(for more detail, see Appendix \ref{ap:prompts}): 

\begin{center}
\textit{<Essay prompt> + Do you agree or disagree? Use specific reasons and examples to support your answer. Write an essay of roughly 300/400/500 words.}
\end{center}

To balance the length of machine essays with human essays, the prompts for WECCL, TOEFL11 and GRE differ in their requirement of essay length (300, 400 and 500 words respectively).

\subsubsection{Generation configuration}

We experiment with essay generation using 7 GPT models (see Table \ref{tab:gen:info}). 
We use all 7 models to generate essays in response to prompts in TOEFL11 and WECCL \footnote{We give gpt2-xl beginning sentences randomly chosen from human essays for continuous writing, and remove those beginning sentences after generation.}. 
However, as our GRE essays are mostly sample essays with high scores,
we generate all 590 GRE machine essays using the two more powerful models:
text-davinci-003 and gpt-3.5-turbo.

For the balance of the ArguGPT corpus, we generate 210 essays for each TOEFL11 prompt (30 essays per model with 6 essays per temperature for \texttt{temp} $\in \{0.2, 0.35, 0.5, 0.65, 0.8\}$), 35-210 essays for each WECCL prompt (the number of human essays for each WECCL prompt is different), and only 1 essay for each GRE prompt.

In our pilot study, we find that GPT-generated essays may have the following three problems: 1) \textbf{Short:} Essays contain only one or two sentences.
2) \textbf{Repetitive:} One essay contains repetitive sentences or paragraphs.
3) \textbf{Overlapped:} Essays generated by the same model may overlap with each other.

Thus we filter out essays with any of the three problems. First, we remove essays shorter than 100 words\footnote{The minimal length of gpt2-xl is set to 50, for it is more difficult for gpt2-xl to generate longer texts.}. 
Then we compute the similarity of each sentence pair in an essay by comparing how many words co-occur in both sentences. 
If 80\% of words co-occur, then the sentence pair is considered to be two \textbf{similar sentences}. 
If 40\% of sentences in one essay are \textbf{similar sentences}, then the essay is considered to be \textbf{repetitive} and will be removed. 
In like manner, we pair sentences in \textit{Essay A} with sentences in \textit{Essay B}. 
If 40\% of sentences in \textit{Essay A} and \textit{Essay B} altogether are \textbf{similar}, then \textit{Essay A} is considered to be \textbf{overlapped} with \textit{Essay B}, which will result in the removal of \textit{Essay B}.

The proportion of essays generated is given in Table \ref{tab:gen:info}. We have generated 9,647 essays in total, with 4,708 valid. 
Then we sample essays from machine-written WECCL/GRE essays to match the number of essays in human-written WECCL/GRE essays. We also manually remove some gpt2-generated essays that are apparently not in the style of argumentative writing (See Appendix \ref{ap:gpt2:deleted}), resulting in 4,038 machine-generated essays in total.

\begin{table}[t]
    \centering
    \resizebox{\textwidth}{!}{
        \begin{tabular}{ccccccc}
            \toprule
            Model & Time stamp & \# total & \# valid & \# short & \# repetitive & \# overlapped \\\midrule
            gpt2-xl & Nov, 2019 & 4,573 & 563 & 1,637 & 0 & 2,373 \\
            text-babbage-001 & April, 2022 & 917 & 479 & 181 & 240 & 17 \\
            text-curie-001 & April, 2022 & 654 & 498 & 15 & 110 & 31 \\
            text-davinci-001 & April, 2022 & 632 & 493 & 1 & 41 & 97 \\
            text-davinci-002 & April, 2022 & 621 & 495 & 1 & 56 & 69 \\
            text-davinci-003 & Nov, 2022 & 1,130 & 1,090 & 0 & 30 & 10 \\
            gpt-3.5-turbo & Mar, 2023 & 1,122 & 1,090 & 0 & 4 & 28 \\\midrule
            \# total & - & 9,647 & 4,708 & 1,835 & 481 & 2,625
            \\\bottomrule
        \end{tabular}
    }
    \caption{Information about machine essays generation. Timestamps (except gpt2-xl) are retrieved from OpenAI's API. }
    \label{tab:gen:info}
\end{table}

\subsection{Preprocessing}

We preprocess all human and machine texts in the same manner, so that the GPT/human-authored texts would not be recognized based on superficial features such as spaces after punctuation and inconsistent paragraph breaks. 

Specifically, we perform the following preprocessing steps.

\begin{itemize}\setlength\itemsep{.1em}
    \item Essays generated by gpt-3.5-turbo often begin with ``As an AI model...'', which gives away its author. Therefore, we remove sentences beginning with ``As an AI model ...''.
    \item There are incorrect uses of capitalization. We capitalize the first letter of every sentence and the pronoun ``I''.
    \item There are incorrect uses of spaces and line breaks. We normalize the use of spaces and line breaks. One space is inserted after every punctuation; one space is inserted between two words; all spaces at the beginning of each paragraph are deleted; two line breaks are inserted at the end of each paragraph.
    \item We normalize the use of apostrophes (e.g., don-t -> don't).
\end{itemize}

\subsection{Collecting out-of-distribution data}

An out-of-distribution (OOD) test set is collected to evaluate the generalization ability of the detectors trained on the in-distribution dataset. 
Ideally, we should pair human and machine essays with the same writing prompts to compose the OOD test set, with which we can see the performance of detectors on both positive and negative samples at the same time. 
However, after compiling the ArguGPT dataset, we find no more human-written argumentative essays with accessible writing prompts. 
Therefore, we simply collect 500 human essays and 500 machine essays respectively without pairing them together. 

The OOD test set is divided into two parts, as Machine OOD and Human OOD.
Machine essays are generated by LLMs and human essays are written by Chinese English learners (see Table~\ref{tab:ood:dataset}). Human essays and machine ones in OOD dataset share different writing prompts. 
In the two independent sub-sets, we can evaluate the performance on negative and positive samples respectively.

\begin{table}[]
    \centering
    \resizebox{\textwidth}{!}{
    \begin{tabular}{ccc||ccc}
        \toprule
        \multicolumn{3}{c}{OOD\_{machine}} & \multicolumn{3}{c}{OOD\_{human}} \\\cmidrule(lr){1-3} \cmidrule(lr){4-6}
        sub-corpus & \# essays & \# tokens & sub-corpus & \# essays & \# tokens \\\midrule
        gpt-3.5-turbo & 100 & 44,028 & st2: high school students & 100 & 19,975 \\
        gpt-4 & 100 & 43,986 & st3: junior college students & 100 & 16,318 \\
        claude-instant & 100 & 31,815 & st4: senior college students & 100 & 17,165 \\
        bloomz-7b & 100 & 29,659 & st5: junior English majors & 100 & 24,978 \\
        flan-t5-11b & 100 & 30,632 & st6: senior English majors & 100 & 54,466  \\\bottomrule
    \end{tabular}
    }
    \caption{Information about the out-of-distribution (OOD) test set}
    \label{tab:ood:dataset}
\end{table}

\paragraph{Human OOD essays} The human OOD dataset is collected to test how the detectors trained with limited writing prompts perform on human essays in response to unseen ones.  
Human essays are sampled from CLEC (Chinese Learner English Corpus)~\cite{clec2003}, containing argumentative essays written by Chinese English learners of five different levels\footnote{Writing prompts of these essays are not published by authors of CLEC.} (see Table \ref{tab:ood:dataset} for details).

Being written by Chinese English learners, essays in CLEC share similar linguistic properties as WECCL. 
However, the writing prompts in CLEC are speculated to be different from WECCL for the topics of these essays never occur in the ArguGPT dataset. 
Therefore, this dataset can be used to evaluate the performance of detectors on the out-of-distribution writing prompts while the linguistic features are probably in-distribution. 

\paragraph{Machine OOD essays} The machine OOD test set is collected to evaluate the performance of the detectors in two cases: (1) new writing prompts, and (2) LLMs that are not used to generate the essays in the training set. 

As the writing prompts in ArguGPT have two parts (i.e., essay prompt and essay prompt), we use the following steps to generate the prompts (see Table~\ref{tab:ood:prompts}): 
half of the writing prompts are composed of 25 unseen essay prompts generated by ChatGPT, with the same added prompt used in the training portion of the ArguGPT dataset as the suffix; 
another half are 25 prompts resulted from combinations of 5 essay prompts sampled from the training set, with 5 unseen added prompts that are again generated by ChatGPT.

\begin{table}[t]
    \centering
    \resizebox{\textwidth}{!}{
    \begin{tabular}{c|c|c|l|p{0.4\textwidth}}
        \toprule
         & type & \# & source & example \\\midrule
        \multirow{7}{*}{\shortstack{w/ unseen\\added prompts}} & \multirow{2}{*}{essay} & \multirow{2}{*}{5} & \multirow{2}{*}{sampled from ArguGPT} & Young people enjoy life more than older people do. \\\cline{2-5}
         & \multirow{5}{*}{added} & \multirow{5}{*}{5} & \multirow{5}{*}{generated by ChatGPT} & Analyze the statement <\textit{essay prompt}> , by examining its causes, effects, and potential solutions. Write an essay of roughly 400 words. \\\hline
        \multirow{8}{*}{\shortstack{w/ unseen\\essay prompts}} & \multirow{3}{*}{essay} & \multirow{3}{*}{25} & \multirow{3}{*}{generated by ChatGPT} & Social media has more harmful effects than beneficial effects on society. \\\cline{2-5}
         & \multirow{5}{*}{added} & \multirow{5}{*}{1} & \multirow{5}{*}{the one used in ArguGPT} & <\textit{essay prompt}> Do you agree or disagree? Use specific reasons and examples to support your answer. Write an essay of roughly 400 words. \\\bottomrule
    \end{tabular}}
    \caption{The machine-written OOD test set contains two types of writing prompts: (1) 25 prompts which have unseen added prompts: 5 sampled essay prompts already used in the training portion of ArguGPT, each combined with 5 ChatGPT-generated added prompts, and (2) 25 %
    ChatGPT-generated essay prompts that are disjoint with the prompts in the training set, with the same added prompt used in the training set of ArguGPT. }
    \label{tab:ood:prompts}
\end{table}

We use five models to generate machine-written argumentative essays in response to above 50 prompts, four of which are not used in the generation of the training data, thus serving our purpose to examine the generalization ability of our detectors: 

\begin{itemize}\setlength\itemsep{.1em}
    \item \textbf{gpt-3.5-turbo}: gpt-3.5-turbo is an in-distribution (ID) model used in the ArguGPT dataset. We want to test the generalization ability on the ID model in response to OOD prompts.
    The essays of gpt-3.5-turbo are collected via OpenAI's API.
    \item \textbf{gpt-4}: gpt-4 is an OOD model from the GPT family. Therefore, we can see how the detectors trained on data generated by previous models predict ones written by the latest one. The essays of gpt-4 are accessed via the web-interface.
    \item \textbf{claude-instant}: claude-instant\footnote{\url{https://claude-ai.ai/}} is a large language model developed by Anthropic. With essays generated by claude-instant, we can test how the detectors transfer the ability to the model from non-GPT family. The essays of claude-instant are collected in the web-interface as gpt-4.
    \item \textbf{bloomz-7b}~\citep{workshop2023bloom} and \textbf{flan-t5-11b}~\citep{2022FLAN}: We also use two language models in a much smaller scale, to investigate how well these detectors can detect essays generated by smaller language models. 
    For these two models we run them locally on two 24GB RAM GPUs to generate essays. 
\end{itemize}

Each model is asked to generate 2 essays for each prompt, amounting to 500 argumentative essays in total, serving as the machine OOD test set for positive samples.

\section{Human evaluation}
\label{sec:human-eval}

Our first research question is 
whether ESL instructors can identify the texts generated by GPT models. To answer this question, we recruit a total of 43 ESL instructors for two rounds of Turing tests. 
In each round, they are asked to identify which 5 essays are machine-written from 10 randomly sampled TOEFL essays. 
They are also asked to share their observations on the linguistic and stylistic characteristics of GPT-generated essays. 
In Section~\ref{sec:human:eval:methods}, we describe details of this experiment. In Section \ref{sec:human:res} we present and analyze the results.

\subsection{Methods}\label{sec:human:eval:methods}

\paragraph{Task}
We ask human participants to determine whether an essay is written by a human or a machine. 
Previous research show that it is difficult for a layperson to spot a machine generated text~\citep{brown2020language-gpt3,dou_is_2022,guo2023close}. 
In light of such discoveries, we present 5 machine generated essays and 5 human essays each round to the participant, and ask them to rate the probability of each text being written by a human/machine on a 6-point Likert Scale, where 1 corresponds to ``definitely human'' and 6 corresponds to ``definitely machine''. 
For the 5 machine essays, we sample  1 from each of the following 5 models: gpt2-xl, text-babbage-001, text-curie-001, text-davinci-003, gpt-3.5-turbo.
For the 5 human essays, we sample 1 low, 3 medium and 1 high level essays, disregarding the native language of the human author.

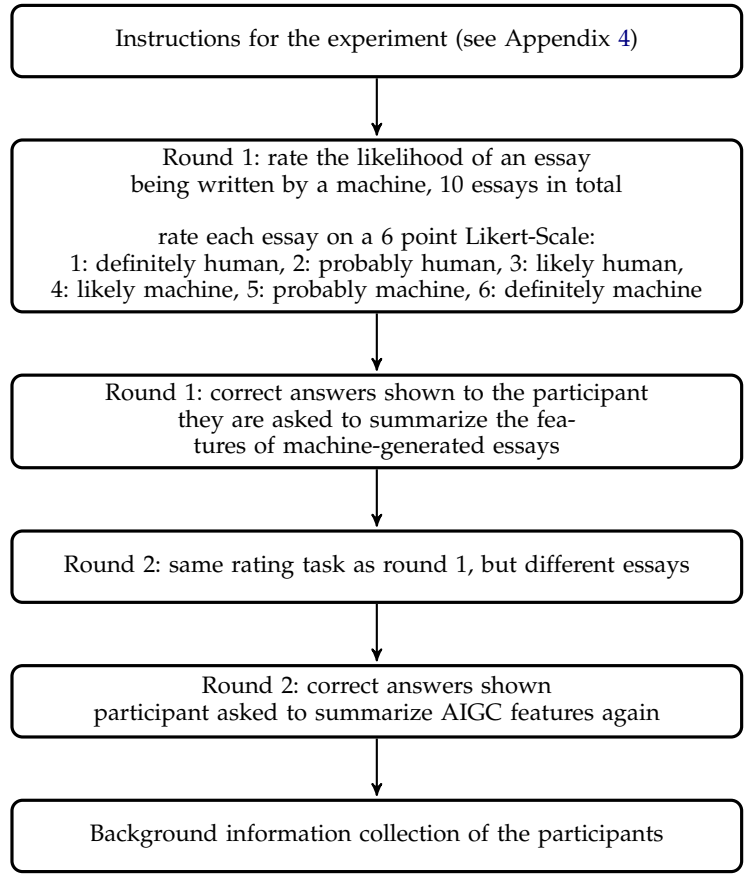
\begin{figure}
    \centering
\small
\begin{tikzpicture}
  [node distance=.8cm,
  start chain=going below]
     \node[punktchain, join] (intro) {Instructions for the experiment (see Appendix~\ref{app:instructions:human:exp})};
     \node[punktchain, join] (probf) {Round 1: rate the likelihood of an essay\\ being written by a machine, 10 essays in total \\ ~ \\ rate each essay on a 6 point Likert-Scale: \\ 1: definitely human, 2: probably human, 3: likely human,\\ 4: likely machine, 5: probably machine, 6: definitely machine};
     \node[punktchain, join] (investeringer)      {Round 1: correct answers shown to the participant \\ they are asked to summarize the features of machine-generated essays};
     \node[punktchain, join] (perfekt) {Round 2: same rating task as round 1, but different essays};
     \node[punktchain, join] (emperi) {Round 2: correct answers shown \\ participant asked to summarize AIGC features again};
      \node [punktchain, join]  {Background information collection of the participants};
  \end{tikzpicture}

    \caption{Pipeline for AIGC identification experiment}
    \label{fig:human:exp:pipeline}
\end{figure}

Each participant will perform such a rating task for two rounds, on two different sets of essays. %
Answers given by participants are correct when they rating 1-3 point for human essays and 4-6 point for machine essays. 
After each round, they will be
presented with the correct answers, giving them a chance to observe the features of the GPT-generated essays, which they are asked to write down and submit in a text box.
Then they will be presented with the next set of essays. See Figure~\ref{fig:human:exp:pipeline} for the pipeline of the experiment. 
We expect the accuracy to be higher in the second set of essays as the participants have seen machine essays and the correct answers in the first round. 
Instructions for the experiment can be found in Appendix \ref{app:instructions:human:exp}.

\paragraph{Participants}

We recruit a total of 43 ESL instructors/teaching assistants from over 5 universities across China. The instructors from the English Department and/or College English Department\footnote{The former is responsible for teaching English majors, while the latter is responsible for teaching general English courses to non-English majors in the university.  } include assistant professors/lecturers, associate professors, professors, and Ph.D. and MA students who have experiences as teaching assistants. Details are presented in Table~\ref{tab:part:id}. 
Each participant are compensated RMB (40 + 2$\times$correct answers), as an incentive for them to try their best in the task.
The mean time of completion for round 1 and 2 is 15 minutes and 10 minutes respectively\footnote{We did not see a strong correlation between completion time and the participants' accuracy in the rating task(Pearson $r$ around 0.1)}.

\begin{table}[t]
\centering
\begin{tabular}{ccc}
\toprule
Identity & \# Participants & Accuracy \\\midrule
MA student & 4 & 0.5875 \\
Ph.D.~Student & 16 & 0.6656 \\
Assi.~Professor/Lecturer & 11 & 0.6364\\
Asso.~Professor & 7 & 0.6929\\
Professor & 3 & 0.6500\\
Other & 2 & 0.5000 \\\midrule
total & 43 & - \\\bottomrule
\end{tabular}
\caption{Proportion of participants' identities and their performance on the AIGC detection experiment.}
\label{tab:part:id}
\end{table}

\subsection{Results}
\label{sec:human:res}
The \textbf{43} participants make \textbf{860} ratings in response to \textbf{280} essays, 
which are taken from the 300 TOEFL essays in the test split of the corpus (see Section~\ref{sec:classifier}). 
We count the number of correct answers among the 860 choices in order to obtain the accuracy.

\subsubsection{Quantitative analysis}
The accuracy of our human participants in identifying machine essays is presented in Table \ref{tab:human:res}. 
From the left side of Table \ref{tab:human:res}, we see that \textbf{the mean accuracy from all subjects in both rounds is \textbf{0.6465}}, roughly 15 percent more than the baseline, which is 0.5, since we have an equal number of human and machine texts in the test set.

One interesting discovery is that \textbf{it is much easier for our participants to identify human essays.} The accuracy of identifying human essays reaches \textbf{0.7744}, while the accuracy of machine is only at chance level: \textbf{0.5186}. 
We believe this is because all of our participants have a lot of experience reading ESOL learners' writings and are thus quite familiar with the style and errors one can find in an essay written by a human language learner.

However, only 11 out of the 43 participants indicated that they are familiar with the (Chat)GPT models; that is, most of them are unfamiliar with the type of text generated by these models, which could explain why they have lower accuracy when identifying GPT-authored texts.

Going down the left side of Table~\ref{tab:human:res}, participants who self-report that they have some familiarity with LLMs have better performance on our task than those who are not familiar with LLMs (0.69 vs 0.64).

We also observe some interesting trends, as shown on the right side of Table~\ref{tab:human:res}. \textbf{Participants are better at identifying \textit{low} level human essays} (acc: 0.8372), and  \textbf{essays generated by more \textit{advanced} models} such as text-davinci-003 and gpt-3.5-turbo (acc: 0.6279). 
Participants are particularly bad at identifying essays generated by gpt2-xl (acc: 0.3721). 
This is different from \citet[][section 1]{clark_all_2021} who suggested that the evaluators ``\textbf{underestimated} the quality of text current models are capable of generating''. When our experiments were conducted, ChatGPT has become the latest model and participants seem to \textbf{overestimate} the non-ChatGPT models, by assigning gpt2-xl essays to human essays, as they have commonly seen student essays with low quality. 

\begin{table}[t]
\centering
\resizebox{\textwidth}{!}{
\begin{tabular}{ccc||ccc}
\toprule
Group by & Group & Accuracy & & Author & Accuracy \\\midrule
\multirow{3}{*}{Essay type} & Overall & 0.6465 & \multirow{5}{*}{M} & gpt2-xl & 0.3721 \\
 & Human essays & 0.7744 & & text-babbage-001 & 0.4651 \\
 & Machine essays & 0.5186 & & text-curie-001 & 0.4651 \\\cline{1-3}
Same essay prompt & Yes & 0.6472 & & text-davinci-003 & 0.6628 \\
for 10 essays & No & 0.6460 & & gpt-3.5-turbo & 0.6279  \\\hline
\multirow{2}{*}{Familiarity} & Not familiar (600 ratings) & 0.6400 & \multirow{3}{*}{H} & human-low & 0.8372 \\
\multirow{2}{*}{w/ GPT} & Familiar (220 ratings) & 0.6909 & & human-medium & 0.7752 \\
 & Other (40 ratings) & 0.5000 && human-high & 0.7093 \\\bottomrule

\end{tabular}}
\caption{Accuracy of participants in the machine/human essay identification experiment. Left: accuracy break-down by essay type, same essay prompt or not, and familiarity w/ GPT. Right: accuracy break-down by essay author. }
\label{tab:human:res}
\end{table}

\begin{table}[t]
\centering
\begin{tabular}{cccc||ccc}
\toprule
 & \multicolumn{3}{c}{Round 1} & \multicolumn{3}{c}{Round 2} \\\cmidrule(lr){2-4} \cmidrule(lr){5-7}
 & Overall & Human & Machine & Overall & Human & Machine \\\midrule
Accuracy & 0.6163 & 0.7535 & 0.4791 & 0.6767 & 0.7954 & 0.5581 \\\bottomrule
\end{tabular}
\caption{Participants' accuracy in two rounds. After round 1, they were given the correct answers and asked to summarize the characteristics of machine text, which serves as some minimal self-training. They evaluated disjoint sets of 10 essays (5 human; 5 machine) in each round.}
\label{tab:rnd:cmp}
\end{table}

\textbf{Subjects' overall accuracy in the first round is \textbf{0.6163}, while performance in the second round rises to \textbf{0.6767}} (see Table~\ref{tab:rnd:cmp}).
This suggests that after some exposure to machine texts (i.e., 5 machine texts and 5 human texts side by side) and reflection on the linguistic features of machine texts, our subjects become 
better at identifying machine texts, with the accuracy rising from 0.4791 to 0.5581. 
This result is in line with \citet{clark_all_2021} who employed 3 methods that ultimately improved human evaluators' judgment accuracy from chance level to 55\%. We also see a 4\% improvement in the accuracy of identifying our human-written essays.

\subsubsection{Qualitative analysis}

In this section, we summarize the features of machine-essays provided by our participants in the experiment. 

First, participants associate two distinctive features with human essays: (1) human essays have more \textbf{spelling and grammatical errors}, which is also mentioned in \citet{dou_is_2022}, and (2) human essays may contain \textbf{more personal experience}. Seeing typos/grammatical mistakes and personal experience in one essay, participants are very likely to 
categorize it as 
human-authored.

As for machine essays, participants generally think (1) machine essays provide many \textbf{similar examples}, and (2) machine essays have \textbf{repetitive expressions}. These two features corroborate the findings of \citet{dou_is_2022}. Reasons when participants make the right choices are presented in Table \ref{tab:correct:example}.

\begin{table}[ht]
    \centering
    \resizebox{\textwidth}{!}{
    \begin{tabular}{p{0.5\textwidth}|c|c|c}
        \toprule
        Text Excerpt & Author & Choice & Reason \\\midrule
        So to the \textbf{oppsite} of the point that mentioned in the theme, I think there will more people choose cars as their first \textbf{transpotation} when they are out and certainly there will be more cars in twenty years. & \multirow{5}{*}{\shortstack{human-\\medium}} & \multirow{5}{*}{Human} & \multirow{5}{*}{\shortstack{There are too many typos\\and grammatical errors.}} \\\hline
        Apart from that the civil service is the \textbf{Germnan} alternative to the militarz service. For the period of one year young people can help in there communities. & \multirow{4}{*}{\shortstack{human-\\high}} & \multirow{4}{*}{Human} & \multirow{4}{*}{\shortstack{The essay might be written\\by a German speaker.}}  \\\hline
        Firstly... when I traveled to Japan... Secondly... when I went on a group tour to Europe... Thirdly... when I went on a safari in Africa... & \multirow{4}{*}{\shortstack{gpt-\\3.5-\\turbo}} & \multirow{4}{*}{Machine} & \multirow{4}{*}{\shortstack{Examples provided\\are redundant.}} \\\hline
        I wholeheartedly... getting a more personalized experience... Some of the benefits... getting a more personalized experience... So, overall... get a more personalized experience...  & \multirow{5}{*}{\shortstack{text-\\curie-\\001}} & \multirow{5}{*}{Machine} & \multirow{5}{*}{\shortstack{There are too many\\repetitive expressions.}}
        \\\bottomrule
    \end{tabular}}
    \caption{Reasons given by participants when their choices are correct. Participants usually rely on reasons presented to identify whether an essay is human-written or machine-generated, especially for identifying human essays according to \textbf{typos} and \textbf{grammatical errors}.}
    \label{tab:correct:example}
\end{table}

However, even with the knowledge of the two features of machine essays discussed above, participants are not very confident when identifying machine essays. 
There are redundant and repetitive expressions in human writings as well, which might confuse participants in this regard. Another important feature frequently mentioned by participants is \textbf{off-prompt}, meaning the writing digresses from the topic given by the prompt.
Some tend to think it is a feature of machine essays, while others think it features human essays. 

After being presented with correct answers in the human evaluation experiment,
participants summarize their impression on AI-generated argumentative essays. 
We list some of the commonly mentioned ones below. 

\begin{itemize}
\setlength\itemsep{.1em}
    \item Language of machines is more fluent and precise. (1) There are no grammatical mistakes or typos in machine-generated essays. (2) Sentences produced by machines have more complex syntactic structures. (3) The structure of argumentation in machine essays is complete.
    \item Machine-generated argumentative essays avoid subjectivity. (1) Machines never provide personal experience as examples. (2) Machines seldom use ``I'' or ``I think''. (3) It is impossible to speculate the background information of the author in machine essays.
    \item Machines hardly provide really insightful opinions. (1) Opinions or statements provided by machines are very general, which seldom go into details. (2) Examples in machine-generated essays are comprehensive, but they are plainly listed rather than coherently organized. 
\end{itemize}

After browsing the reasons given by our participants, we find that grammatical mistakes (including typos) and use of personal experience are usually distinctive and effective features to identify human essays (English learners in our case), 
according to which participants can have a higher accuracy. 
However, a fixed writing format and off-prompt are not so reliable.
If participants identify machine essays by these two features, the accuracy will drop. For a fixed writing format and off-prompt often feature human-authored essays as well.

\subsection{Summary}

Our results indicate that knowing the answers of the first round test is helpful for identifying texts in the second round, which is consistent with \citet{clark_all_2021}. \textit{Contra} \citet{brown2020language-gpt3}, our results show that essays generated by more advanced models are more distinguishable. Moreover, the accuracy on identifying texts generated by models from GPT2 and GPT3 series are lower than what is reported in previous literature~\citep{uchendu2021turingbench, clark_all_2021,brown2020language-gpt3}.
It indicates that \textbf{human participants anticipate that machines are better than human at writing argumentative essays.} When the models (e.g., gpt2-xl) generate an essay of lower quality, our human participants might feel more confused because they expect the machines should have better performance.

Looking into the summaries of AIGC features provided by participants, we can see that participants \textbf{have a more detailed picture of how human essays look like} (e.g., the mistakes that human English learners are likely to make). On the other hand, they can also \textbf{capture some features of machine essays after reading several of them}, though these features were not strong enough to help participants determine whether essays are human- or machine-authored.

Therefore, we think that if English teachers become more familiar with AIGC, they will be more capable to identify the features between human and machine.

\section{Linguistic analysis}
\label{sec:ling-ana}

In this section, we compare the linguistic features of machine and human essays. 
We group the essays by author: (1) low-level human, (2) medium-level human, (3) high-level human, (4) gpt2-xl, (5) text-babbage-001, (6) text-curie-001, (7) text-davinci-001, (8) text-davinci-002, (9) text-davinci-003, and (10) gpt-3.5-turbo.

We first present some descriptive statistics of human and machine essays. We then use established measures and tools in second-language (L2) writing research to analyze the syntactic complexity and lexical richness of both human and machine written texts~\citep{Lu2012TheRO,Lu2010AutomaticAO}.

\subsection{Methods}

\paragraph{Descriptive statistics}

We use in-house Python scripts and NLTK~\citep{Bird2002NLTKTN} to obtain descriptive statistics of the essays in the following 5 measures: (1) mean essay length, (2) mean number of paragraphs per essay, (3) mean paragraph length, (4) mean number of sentences per essay, and (5) mean sentence length.

\paragraph{Syntactic complexity}

To analyze the syntactic complexity of the essays, we apply the L2 Syntactic Complexity Analyzer to calculate 14 syntactic complexity indices for each text~\citep{Lu2010AutomaticAO}. 
These measures have been widely used in L2 writing research. 
Table \ref{tab:syntactic:definition} presents details of the indices.
However, only six out of these 14 measures are reported by~\citet{Lu2010AutomaticAO} to be linearly correlated with language proficiency levels. Therefore, we only present the results for these six measures.

\begin{table}[t]
\centering
\small
\resizebox{\textwidth}{!}{
\begin{tabular}{lllll}
\toprule
Measure & Code & Definition & & \\
\midrule
\multicolumn{3}{l}{Length of production unit} & &  \\ \midrule
Mean length of clause & MLC & \# of words / \# of clauses &  &  \\
Mean length of T-unit & MLT  & \# of words / \# of T-units & & \\
\midrule
\multicolumn{3}{l}{Coordination} & &  \\ \midrule
Coordinate phrases per clause & CP/C & \# of coordinate phrases / \# of clauses & &  \\
Coordinate phrases per T-unit & CP/T & \# of coordinate phrases / \# of T-units & &  \\
\midrule
\multicolumn{3}{l}{Particular structures} & & \\ \midrule
Complex nominals per clause & CN/C & \# of complex nominals / \# of clauses & &  \\
Complex nominals per T-unit & CN/T & \# of complex nominals / \# of T-units & &  \\
\bottomrule
\end{tabular}
}
\caption{The six syntactic complexity measures used in this study, replicated from Table 1 of~\citet{Lu2010AutomaticAO}. A clause here is a syntactic structure with a subject and a finite verb~\citep{hunt1965clause}. A T-unit here is one main clause with or without subordinate clauses or nonclausal structure ~\citep{hunt1970tunit}. These 6 measures are chosen from 14 measures from \citet{Lu2010AutomaticAO} for analysis since he reported them to be positively related to proficiency level with statistically significant between-level differences.  
}\label{tab:syntactic:definition}
\end{table}

\paragraph{Lexical complexity}

Lexical complexity or richness is a good indicator of essay quality and  
has been considered a useful and reliable measure to examine the proficiency of L2 learners~\citep{laufer1995vocabulary}. 
Many L2 studies have discussed the criteria for evaluating lexical complexity. 
In this study, we follow \citet{Lu2012TheRO} who compared 26 measures in language acquisition literature and developed a computational system to calculate lexical richness from three dimensions: lexical density, lexical sophistication, and lexical variation.

Lexical density means the ratio of the number of lexical words to the total words in a text. We follow \citet{Lu2012TheRO} to define lexical words as nouns, adjectives, verbs, as well as adverbs with an adjectival base, such as ``well" and the words formed by ``-ly" suffix. 
Modal verbs and auxiliary verbs are not included. 
Lexical sophistication calculates the advanced or unusual words in a text~\citep{read2000assessing}. We further operationalize sophisticated words as the lexical words, and verbs which are not on the list of the 2,000 most frequent words generated from the American National Corpus. 
As for lexical variation, it refers to the use of different words and reflects the learner's vocabulary size. 

Concretely, we apply the Lexical Complexity Analyzer from \citet{Lu2012TheRO} to calculate 25 metrics for each essay. See Table \ref{tab:lexical:measures} in Appendix \ref{sec:app-ling-analysis} for details of the measures and formulas. We use spaCy~\citep{honnibal-johnson:2015:EMNLP} to perform tokenization and POS-tagging of all essays.

\paragraph{N-gram analysis}

We use the NLTK package to extract trigrams, 4-grams and 5-grams from both human and machine essays and calculate their frequencies in order to find out the  usage preferences in human and machine essays. 
We then compute log-likelihood~\citep{rayson2000comparing} for each N-gram in order to uncover phrases that are overused in either machine or human essays. 

\subsection{Results}

\subsubsection{Descriptive statistics}

\begin{figure}[t]
    \centering
    \includegraphics[width=\textwidth]{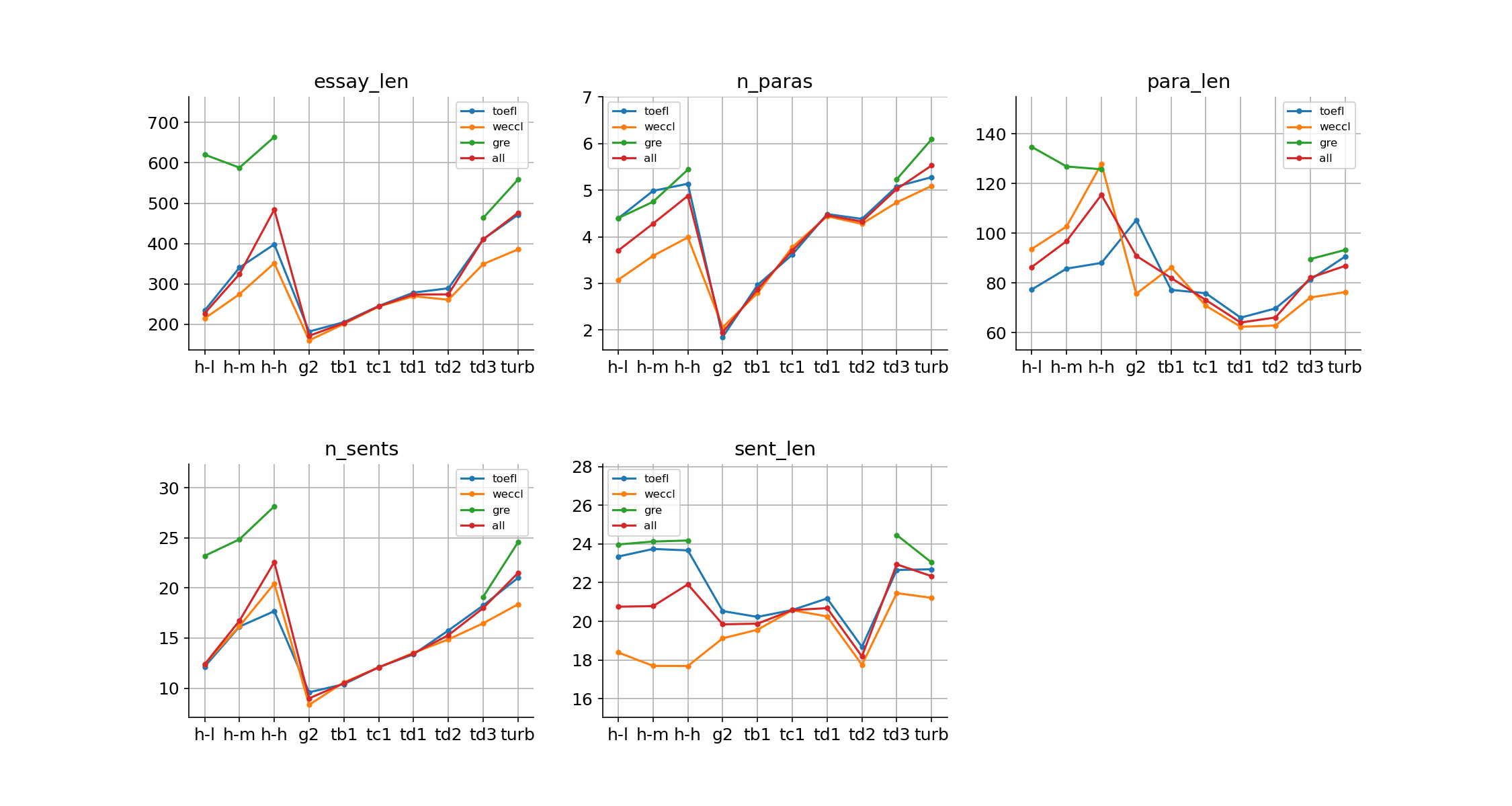}
    \caption{Descriptive statistics of sub-corpora in the ArguGPT corpus. We show mean length in words of the essays (essay\_len), paragraphs (para\_len), and sentences (sent\_len), and mean number of paragraphs (n\_paras) and sentences in one essay (n\_sents). Labels in the x-axis: h-l, h-m and h-h: human-low/medium/high; g2: gpt2-xl; tb1: text-babbage-001; tc1: text-curie-001; td1/2/3: text-davinci-001/002/003; turb: gpt-3.5-turbo. Same below. }
    \label{fig:ling:desc}
\end{figure}

The descriptive statistics of sub-corpora in the ArguGPT corpus are presented in Figure \ref{fig:ling:desc}. As for humans, essays with higher scores are likely to have a longer length from both essay-level and paragraph-level and more paragraphs and sentences in one essay. However, humans of all three levels write sentences of similar length.

In like manner, more advanced AI models 
are likely to have longer essay length, more paragraphs and sentences. However, the mean length of paragraphs goes down from gpt2-xl to text-davinci-001, and goes up slightly from text-davinci-002 to gpt-3.5-turbo. As for the mean length of sentences, text-davinci-002 writes the shortest sentences among machines.

Different machines match humans of different performance levels of mean essay-, paragraph-, and sentence-level length. However, regarding the length of paragraphs, human writers outperform machine writers.

\subsubsection{Syntactic complexity}

\begin{figure}[t]
    \centering   
    \scalebox{1}{
    \includegraphics[width=1.0\textwidth]{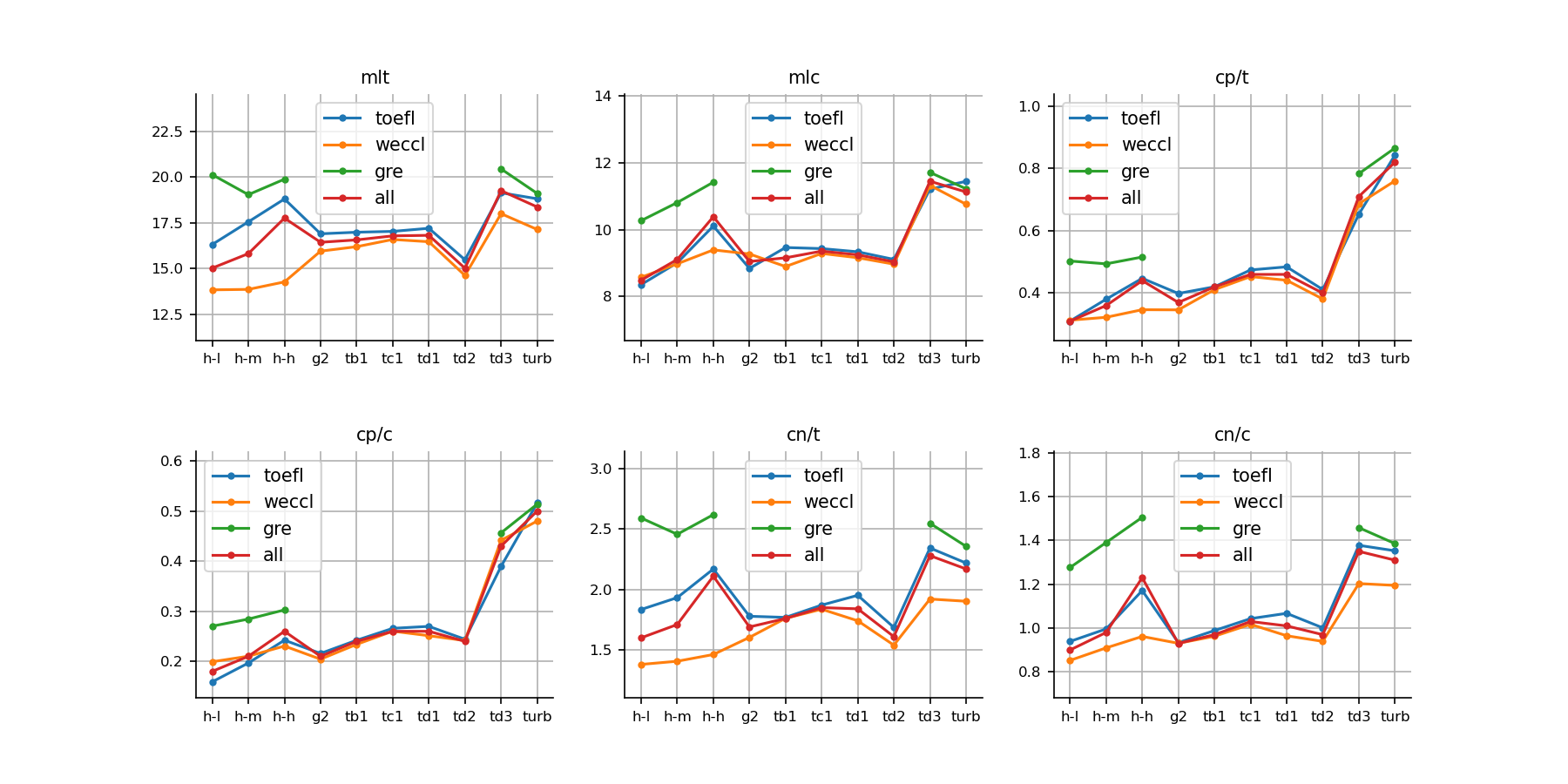}
    }
    \caption{6 measures of syntactic complexity, obtained by using the tool from \citet{Lu2010AutomaticAO}. mlt: mean length of T-unit; mlc: mean length of clause; cp/t and cp/c: \# coordinating phrases per T-unit/clause; cn/t and cn/c: \# of complex nominals per T-unit/clause. }
    \label{fig:ling:syn:complexity}
\end{figure}

Figure~\ref{fig:ling:syn:complexity} gives the means of the six syntactic complexity values of the essays which are grouped by sub-corpora in the ArguGPT corpus.

As shown in Figure~\ref{fig:ling:syn:complexity}, all 6 chosen syntactic complexity values progress linearly across 3 score levels for human essays. They also indicate a general growing trend across the language models following the order of development. It is noticeable that text-davinci-002 is worse w.r.t.~these measures than previous and later models. 
According to MLT (Mean length of T-unit) and CN/T (complex nominals per T-unit), it is even outperformed by gpt2-xl. 

When we compare human essays with machine essays, even high-level human English learners are outperformed by gpt-3.5-turbo and text-davinci-003 in all 6 measures.
This is particularly true for CP/T and CP/C, which indicates that coordinate phrases are much more common in the essays from the last two models than from human learners.
On the other hand, gpt2, text-babbage-001, text-curie-001, and text-davinci-001/002 seem to be on par with human learners on these measures. 

We take the above results to suggest that in general, more powerful models such as davince-003 and ChatGPT produce syntactically more complex essays than even high-level English learners.

\subsubsection{Lexical complexity}
\begin{figure}[t]
    \centering   
    \scalebox{1}{
    \includegraphics[width=1.0\textwidth]{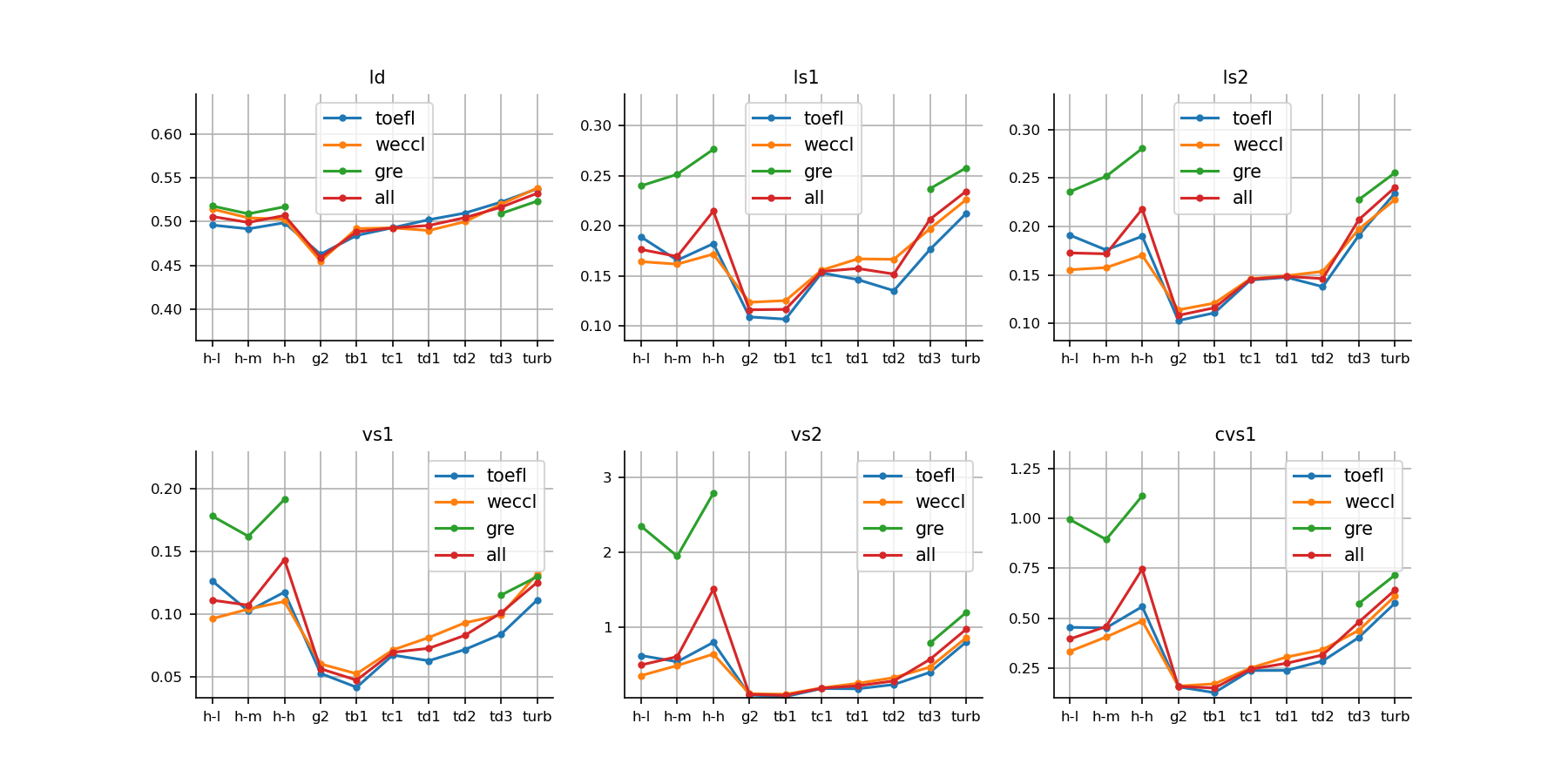}
    }
    \caption{Lexical density and lexical sophistication of ArguGPT corpus. LD refers to lexical density. LS1 and LS2 are two measures of lexical sophistication. VS1 and VS2 are two different ways to calculate verb sophistication. CVS1 is the corrected version of VS1. }
    \label{fig:ling:lexi1}
\end{figure}

\begin{figure}[t]
    \centering   
    \scalebox{1}{
    \includegraphics[width=1.0\textwidth]{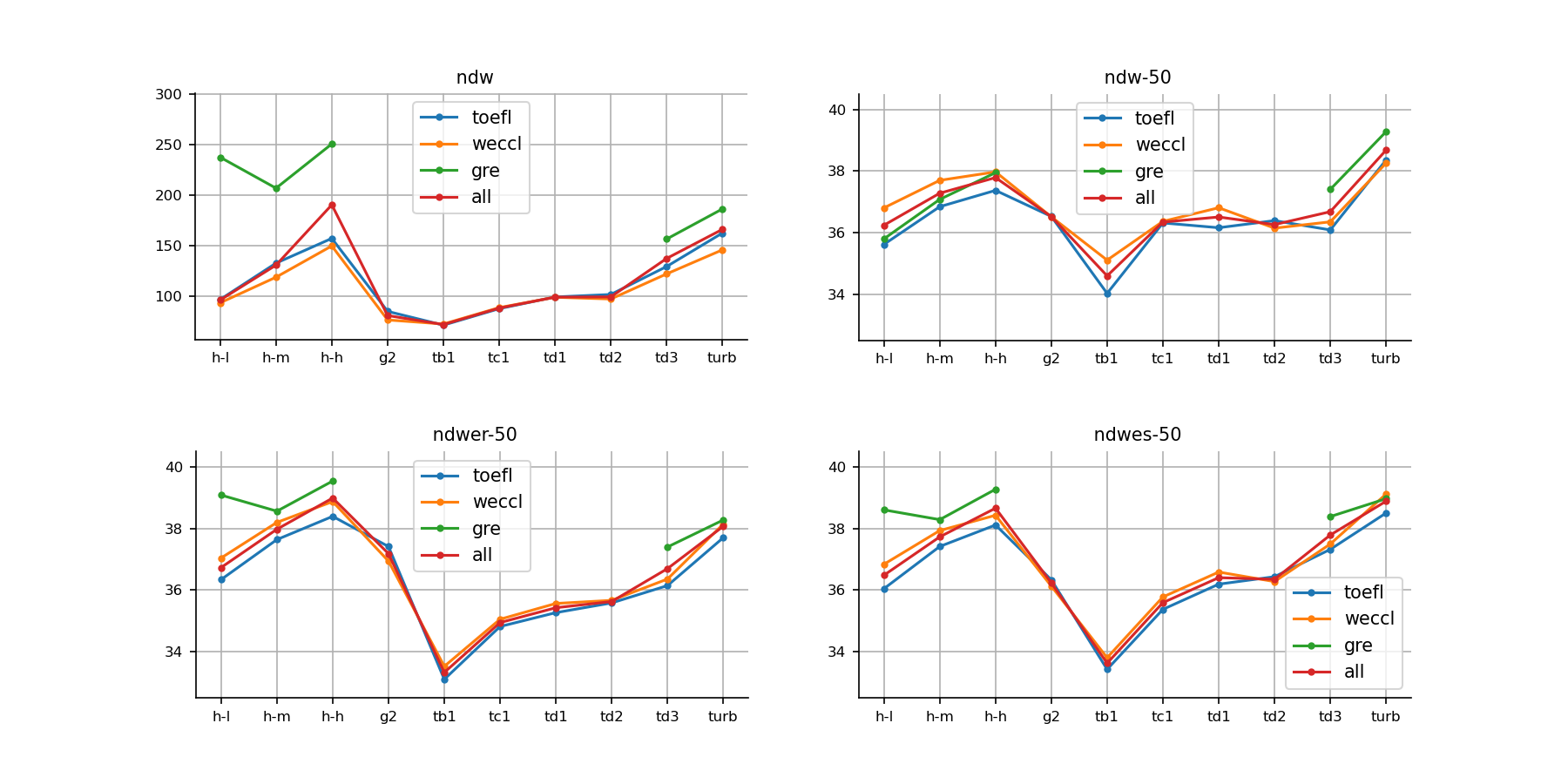}
    }
    \caption{Lexical variation of ArguGPT corpus. NDW stands for the number of different words. The other three indicators are variants of NDW.}
    \label{fig:ling:lexi2}
\end{figure}

\begin{figure}[t]
    \centering    
    \scalebox{1}{
    \includegraphics[width=1.0\textwidth]{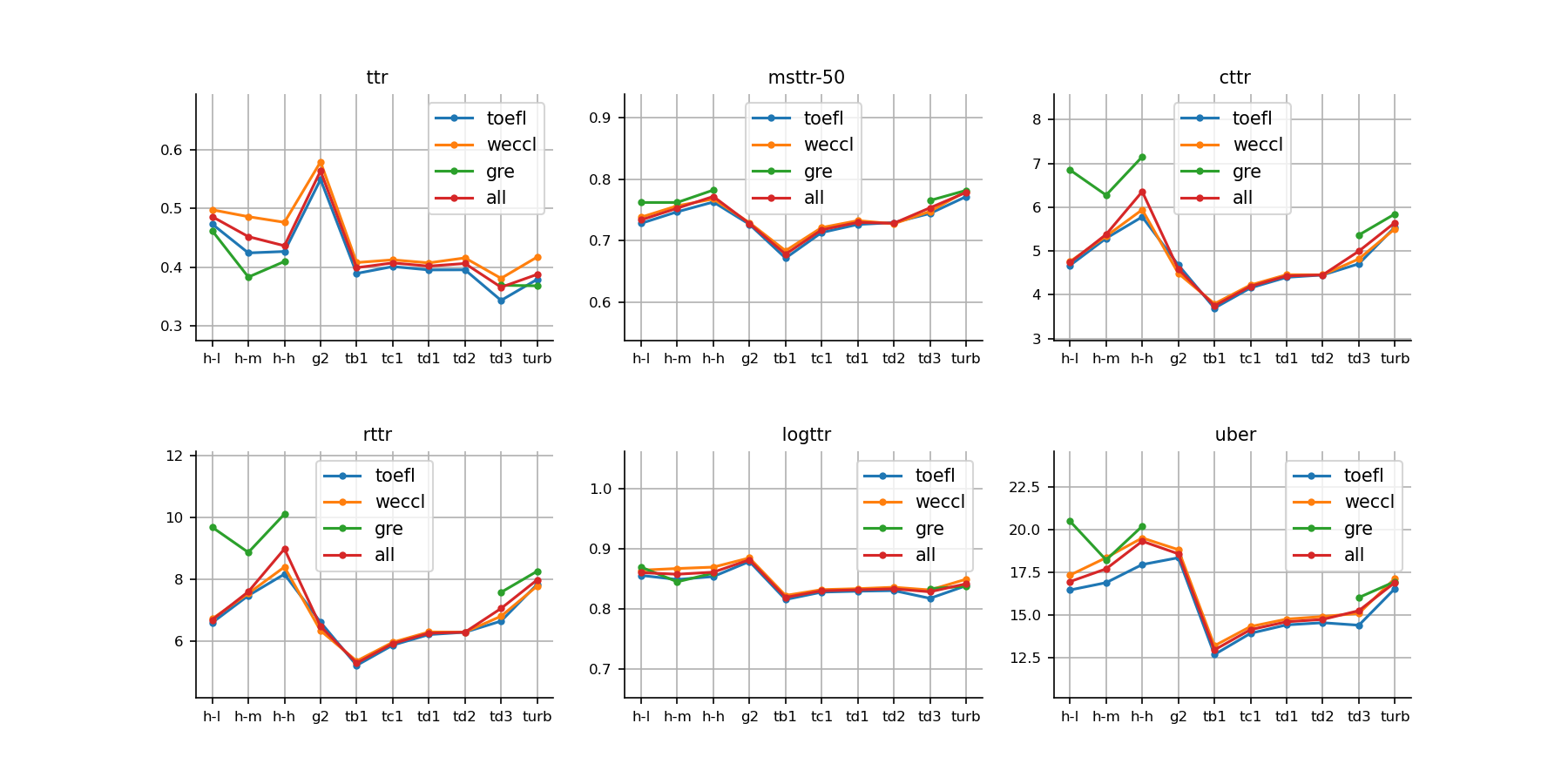}
    }
    \caption{Type-token ratio (ttr) of the essays and five transformations of it. }
    \label{fig:ling:lexi3}
\end{figure}

\begin{figure}[t]
    \centering   
    \scalebox{1}{
    \includegraphics[width=1.0\textwidth]{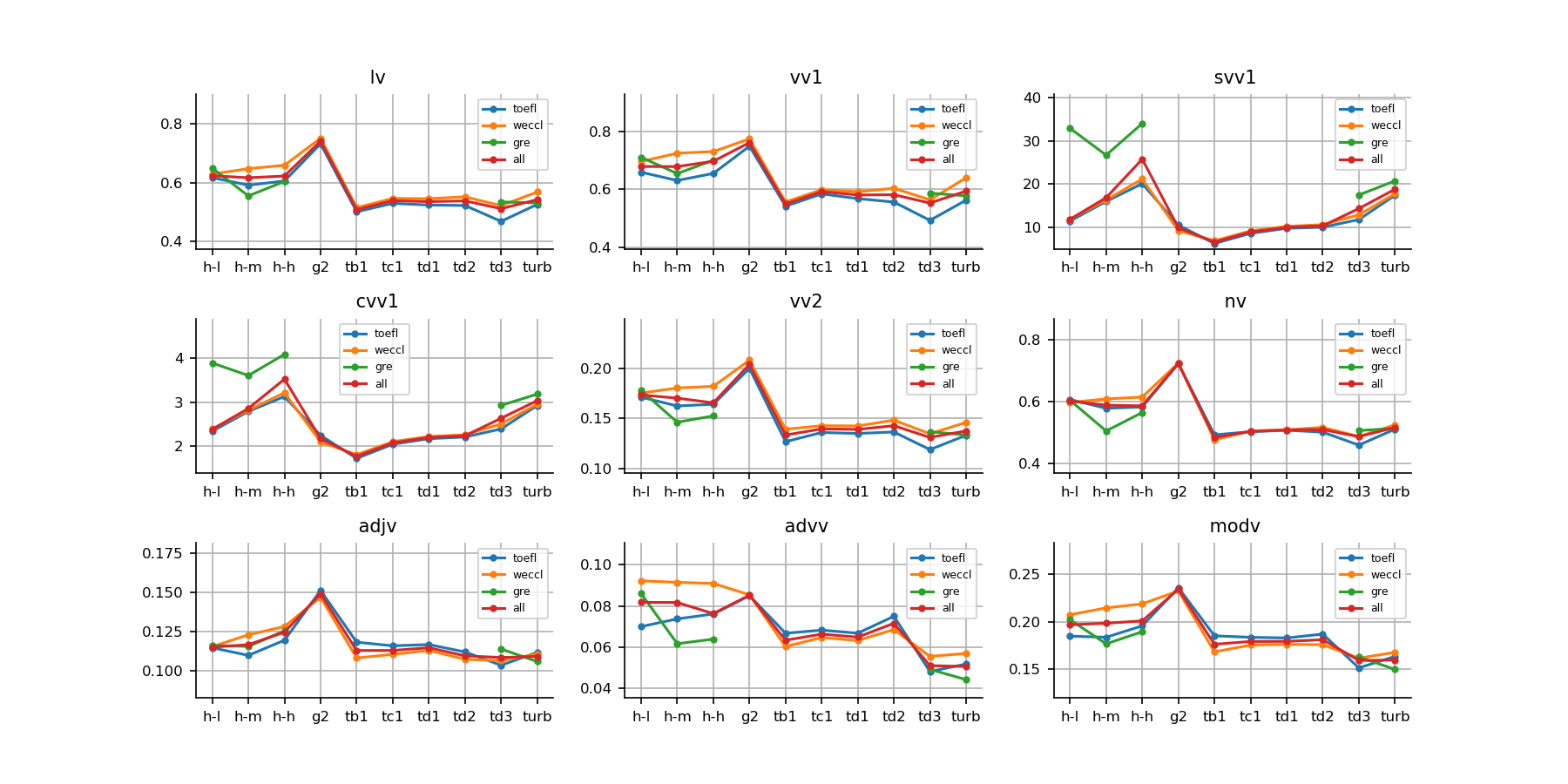}
    }
    \caption{The variation of lexical words and other five ratios, including lexical word variation (lv), verbs (verb variation 1---vv1, squared vv1---svv1, corrected vv1---cvv1 and verb variation 2---vv2), nouns variation (nv), adjective variation (adjv), adverb variation (advv) and modifer variation (modv).}
    \label{fig:ling:lexi4}
\end{figure}

The lexical complexity of  ArguGPT corpus is presented in Figures~\ref{fig:ling:lexi1} to \ref{fig:ling:lexi4} (for actual numbers, see Table \ref{tab:lexical:density:sophistication} and Table \ref{tab:lexical:variation} in the Appendix).

\paragraph{Lexical density}

Lexical density in Figure~\ref{fig:ling:lexi1} shows that human essays tend to use more function words compared to text-davinci-003 and gpt-3.5-turbo, as these two models prefer more lexical words.

\paragraph{Lexical sophistication}

As for lexical sophistication (also shown in Figure~\ref{fig:ling:lexi1}), advanced L2 learners outperform or are on par with gpt-3.5-turbo in all five indicators (lexical sophistication 1/2, and three measures of verb sophistication). 

In terms of verb sophistication (bottom row of Figure~\ref{fig:ling:lexi1}), the differences are pronounced between low/medium level and high level of human writing. Advanced learners surpass gpt-3.5-turbo while the intermediate level is on par with text-davinci-003. However, high-level human essays in WECCL perform worse than gpt-3.5-turbo. Moreover, GRE essays have much higher values than WECCL and TOEFL in these three measures, especially for the advanced level. 
We attribute this to the nature of our GRE corpus, where essays are all example essays for those preparing for the GRE test to emulate rather than representative of all levels of test takers.

\paragraph{Lexical variation}

Measures for lexical variation are shown in Figure~\ref{fig:ling:lexi2} (number of different words), Figure~\ref{fig:ling:lexi3} (type-token ratio) and Figure~\ref{fig:ling:lexi4} (type-token ratio of word class). They indicate the range and diversity of a learner's vocabulary. 

Among the four measures of the number of different words in Figure~\ref{fig:ling:lexi2}, advanced learners exceed gpt-3.5-turbo in two metrics. Text-davinci-003 and gpt-3.5-turbo are comparable to medium and high levels of L2 speakers' writing, respectively.
The trend is similar in WECCL and TOEFL corpora, except for the GRE corpus. Our GRE essays have the largest vocabulary and surpass gpt-3.5-turbo in three metrics. 

Type-token ratio (TTR) is an important measure of lexical richness. Six indices of TTR in Figure~\ref{fig:ling:lexi3} all suggest that, while gpt-3.5-turbo excels the medium-skilled test takers in WECCL and TOEFL, there is still a discernible gap between the performance of machines and that of skillful non-native speakers. GRE test takers at all levels outstrip the machines. 
It should be emphasized that TTR is not standardized compared to CTTR (corrected TTR) and its other variants, and 
shorter essays tend to have a higher TTR. Therefore, essays generated by gpt2-xl have a higher TTR. Other standardized variants better represent the lexical richness of the essays.

Type-token ratio can be further explored with respect to each word class. Figure~\ref{fig:ling:lexi4} shows
the variation of lexical words and other five ratios, including verbs, nouns, adjectives, adverbs and modifiers. We observe that advanced learners outperform gpt-3.5-turbo in all metrics among the three corpora. The margins are obvious in lexical words, verbs, nouns and adverbs.
Note that SVV1 and CVV1 are standardized compared with other metrics, and may be more suitable for analysis of the discrepancies. The verb system is recognized as the focus in second language acquisition, for it is essential to construct any language~\citep{housen2002corpus}, and humans showcase a stronger ability in applying abundant verbs than machines. 

\textcolor{black}{We take these results to suggest that unlike syntactic complexity, high-level English learners and native speakers of English are on par or even exceed gpt-3.5-turbo in terms of lexical complexity.}

\subsubsection{N-gram analysis}
Table \ref{tab:trigram} lists 20 trigrams that are used significantly more frequently by language models than human. It is worth noting that ``i believe that'' appears 2,056 times in 3,338 machine-generated essays, but only 207 times in 3,415 human essays. This seems to be a pet phrase for text-davinci-001, as the phrase occurs 503 times in 509 texts generated by text-davinci-001, while the number of occurrences in 284 essays produced by gpt2-xl is only 31.

\begin{table}[t]
\centering

\resizebox{\textwidth}{!}{
\begin{tabular}{cccc||cccc}
\toprule
\multicolumn{4}{c}{Overused by machines} & \multicolumn{4}{c}{Overused by humans} \\\cmidrule(lr){1-4} \cmidrule(lr){5-8}
Trigram & log-lklhd & M & H & Trigram & log-lklhd & M & H \\ \midrule
i believe that & 1987.2 & 2056 & 207 & more and more & 313.4 & 179 & 753\\
can lead to & 1488.6 & 1152 & 32 & what's more & 230.4 & 2 & 197 \\
more likely to & 1257.1 & 1034 & 43 & the young people & 205.6 & 5 & 193 \\
it is important & 1063.8 & 1130 & 122 & we have to & 194.7 & 29 & 269 \\
are more likely & 831.9 & 679 & 27 & in a word & 184.7 & 1 & 154 \\
be able to & 775.3 & 1296 & 311 & to sum up & 178.7 & 3 & 161\\
is important to & 646.7 & 707 & 82 & most of the & 177.6 & 27 & 247 \\
lead to a & 644.0 & 554 & 29 & in the society & 175.9 & 1 & 147 \\
a sense of & 531.4 & 562 & 60 & and so on & 171.1 & 24 & 232 \\
this can lead & 528.2 & 364 & 2 & we all know & 157.9 & 4 & 149 \\
can help to & 496.6 & 373 & 8 & the famous people & 156.7 & 4 & 148 \\
understanding of the & 493.7 & 507 & 50 & the same time & 156.4 & 48 & 282\\
believe that it & 470.6 & 439 & 32 & of the society & 147.3 & 15 & 182\\
this is because & 468.2 & 564 & 81 & we can not & 147.1 & 43 & 260 \\
likely to be & 459.7 & 422 & 29 & i think the & 144.6 & 17 & 186 \\
this can be & 455.0 & 445 & 38 & as far as & 144.3 & 3 & 133 \\
believe that the & 427.9 & 499 & 67 & so i think & 142.1 & 2 & 126 \\
the world around & 421.2 & 345 & 14 & at the same & 138.4 & 57 & 284 \\
may not be & 410.9 & 504 & 75 & his or her & 133.7 & 19 & 182 \\
skills and knowledge & 404.3 & 292 & 4 & i want to & 133.5 & 13 & 163
\\\bottomrule
\end{tabular}
}

\caption{Top 20 trigrams ranked by log-likelihood~\citep{rayson2000comparing}, either overused by machines (left), or humans (right). Columns ``M'' and ``H'' represent the number of occurrences of the trigram in machine and human essays respectively. }\label{tab:trigram}
\end{table}

The right side of Table~\ref{tab:trigram} lists 20 trigrams which are used significantly more frequently by humans. We can see from the log-likelihood value that the difference in usage of these phrases between humans and machines is not as prominent as those overused by machines. Nevertheless, it is still noticeable that ``more and more'' appears more often in human writing. When looking into the usage of this phrase in the TOEFL corpus, which contains essays written by English learners with 11 different native languages, we found that it is preferred by students whose first language is Chinese or French.

\subsection{Summary}

To sum up, the results of the linguistic analysis suggest that for syntactic complexity, text-davinci-003 and gpt-3.5-turbo produce syntactically more complex essays than most ESOL learners, especially on measures related to coordinating phrases, while other models are on par with the human essays in our corpus. 

For lexical complexity, models like gpt2-xl, text-babbage/curie-001, and text-davinci-001/002, for the most part, are on par with elementary/intermediate English learners on several measures, while text-davinci-003 and gpt-3.5-turbo are approximately on the level of advanced English learners. For some lexical richness measures (verb sophistication, for instance) and GRE essays, our human data show even greater complexity than the best AI models. 

In other words, \textbf{sentences produced by machines are of greater length and syntactic complexity, but words used by machines are likely of less richness and diversity when compared to human writers (mostly English learners in our case).}

Moreover, from the N-gram analysis, machines prefer expressions such as ``I believe that'', ``can lead to'', and ``more likely to'', while these phrases are seldom used in our human-authored essays.

Additionally, we find that there is a general trend of increasing lexical/syntactic complexity as the models progress, but for certain measures, gpt2-xl and text-davinci-002 seem to be the exceptions.

\section{Building and testing AIGC detectors}
\label{sec:classifier}

In this section, we examine whether machine-learning classifiers can distinguish machine-authored essays from human-written ones. 

Specifically, we test two existing AIGC detectors---GPTZero and RoBERTa of \citeauthor{guo2023close}, and build and test our own detectors---SVM models with different linguistic features and a deep learning classifier based on RoBERTa. We also experiment zero/few-shot learning with gpt-3.5-turbo.

\subsection{Experimental settings}

We split the ArguGPT corpus into train, dev, and test sets (see Table \ref{tab:split:data}), where the dev and test sets contain 700 essays respectively. In the dev and test sets, the proportion of low : medium : high level human essays is kept to 1:3:1 for WECCL and TOEFL11. As for WECCL and TOEFL11 machine essays, we sample 150 essays for each exam that are generated by 5 models (excluding text-davinci-001 and text-davinci-002). 
For GRE essays in the dev and test sets, we randomly sample 50 human essays, and 50 machine essays generated by text-davinci-003 and gpt-3.5-turbo. 

For the two existing detectors---GPTZero and the RoBERTa model from~\citep{guo2023close}, we use them as off-the-shelf tools without further fine-tuning or modification. 

We train SVMs and finetune a RoBERTa-large model using the training set of ArguGPT. We perform zero/few-shot learning with gpt-3.5-turbo via OpenAI's API. 

For all experiments, we train and evaluate on document-, paragraph-, and sentence-level.

\begin{table}[t]
    \centering
    \begin{tabular}{c|ccc|ccc}
        \toprule
        Split & \# WECCL & \# TOEFL11 & \# GRE & \# Doc & \# Para & \# Sent \\\midrule
        train & 3,058 & 2,715 & 980 & 6,753 & 29,124 & 111,283 \\
        dev & 300 & 300 & 100 & 700 & 2,947 & 11,302 \\
        test & 300 & 300 & 100 & 700 & 2,953 & 11,704 \\\bottomrule
    \end{tabular}
    \caption{Information of the training, dev and test sets. We split the dataset according to the number of essays. Afterwards, these essays are broken down into paragraphs and sentences for the experiment on detecting texts at different granularities.}
    \label{tab:split:data}
\end{table}

\subsection{Results}

The results for all detectors are summarized in Table \ref{tab:detector:allrestest}. 

\begin{table}[t]
    \centering
    \resizebox{\textwidth}{!}{
    \begin{tabular}{cc||cccc||cc}
    \toprule
    \multicolumn{2}{c}{\textbf{Test set}} & \multicolumn{4}{c}{\textbf{Our detectors}} & \multicolumn{2}{c}{\textbf{Existing detectors}} \\\cmidrule(lr){1-2} \cmidrule(lr){3-6} \cmidrule(lr){7-8}
    \textbf{Test data} & \textbf{maj.~bsln} & \multicolumn{2}{c}{\textbf{Train data}} & \textbf{RoBERTa} & \textbf{Best SVM} & \textbf{GPTZero} & \textbf{\citet{guo2023close}} \\\midrule
    \multirow{4}{*}{Doc} & \multirow{4}{*}{50} & \multirow{4}{*}{Doc} & all & 99.38 & 95.14 & \multirow{4}{*}{96.86} & \multirow{4}{*}{89.86} \\
     & & & 50\% & 99.76 & 94.14 & & \\
     & & & 25\% & 99.14 & 93.86 & & \\
     & & & 10\% & 97.67 & 92.29 & & \\ \hline
    \multirow{2}{*}{Para} & \multirow{2}{*}{52.62} & Doc & \multirow{2}{*}{all} & 74.58 & 83.61 & \multirow{2}{*}{92.11} & \multirow{2}{*}{79.95} \\
     & & Para & & 97.88 & 90.55 & & \\ \hline
    \multirow{2}{*}{Sent} & \multirow{2}{*}{54.18} & Doc & \multirow{2}{*}{all} & 49.73 & 72.15 & \multirow{2}{*}{90.10} & \multirow{2}{*}{71.44} \\
     & & Sent & & 93.84 & 81 & & \\\bottomrule
    \end{tabular}
    }
    \caption{Summary of results for all detectors on the in-distribution test set, evaluated by accuracy. \textcolor{black}{Note that numbers 10\%, 25\%, 50\%, and \textit{all} refer to the percentage of training sets used in ablation studies.} Zero/few-shot ChatGPT is not included in this table, because we don't run all experiments of ChatGPT for its high cost but poor performance.}
    \label{tab:detector:allrestest}
\end{table}

\subsubsection{Existing detectors for AIGC: GPTZero and RoBERTa from~\citep{guo2023close}}

We test two existing detectors for AIGC: GPTZero and finetuned RoBERTa of \citeauthor{guo2023close} \footnote{\url{https://huggingface.co/Hello-SimpleAI/chatgpt-detector-roberta}}.

We evaluate each detector's performance on the ArguGPT test set.
The RoBERTa model from \citet{guo2023close}, which was finetuned on text in other genres such as finance and medicine,  achieves an accuracy of around 90\% for document-level classification, lower than the other classifiers, possibly due to the nature of their training data. 
The same model achieves 79.95\% and 71.44\% respectively on paragraph and sentence level classification. 

As for GPTZero, we use their API, which returns
the probability of being written by an AI model; the returned result includes probabilities for the entire essay, each paragraph and each sentence. 
We consider any text with a probability higher than 0.65 to be AI-written, following the documentation of GPTZero.
Its performance is shown in Table~\ref{tab:detector:allrestest}.
We can see that GPTZero reaches very high accuracy on document, paragraph and sentence levels (all above 90\%).

\subsubsection{SVM detector}

In this section, we aim to find out if AIGC and human-written essays are distinguishable using hand-crafted linguistic features and a SVM classifier. 
By training SVMs on document-level with
syntactic and stylistic features (i.e., no content information) as well as word unigrams (which include content words) and comparing their performance,
\textcolor{black}{we select the best set of features for classification and apply these features on paragraph and sentence levels.}
\paragraph{Linguistic Features}
We select some commonly used linguistic features and extract them with \textit{translationese}\footnote[10]{\url{https://github.com/huhailinguist/translationese} which is extended from \url{https://github.com/lutzky/translationese}.} package, which has been used in linguistically informed text classification tasks for translated and non-translated texts
\citep{volansky2015features,hu2021investigating}. 
The implementation uses Stanford CoreNLP~\citep{Manning2014corenlp} for POS-tagging and constituency parsing. Specifically, we experimented with the following features:

\begin{itemize}
\item \textbf{CFGRs: } The ratio of each  context-free grammar (CFG) rule in the given text to the total number of CFGRs.
\item \textbf{function words: }The normalized frequency of each function word's occurrence in the chunk. Our list of function words comes from~\citep{koppel2011translationese}. It consists of function words and some content words that are crucial in organizing the text, which constitutes 467 words in total. 
\item \textbf{most frequent words: }The normalized frequencies (ratio to the token number) of the most frequent $k$ words ($k=10/50$) in a large reference corpus TOP\_WORDS\footnote[11]{\url{https://github.com/huhailinguist/translationese/blob/master/translationese/word_ranks.py}}. The top words are mainly function words. 
\item \textbf{POS unigrams: }The frequency of each unigram of POS.
\item \textbf{punctuation: }The frequency (ratio to the token number) of punctuation in a given chunk.
\item \textbf{word unigrams: }The frequency of the unigram of words. We only include unigrams with more than three occurrences in the given text.

\end{itemize}

Among the features, \textit{CFGRs} and \textit{POS unigrams} separately represent the sentence structures and part of speech preference. They are the stylistic features reflecting the patterns underlying the superficial language expressions. \textit{function words}, \textit{word unigrams} and \textit{most frequent words} reveal the concrete choices in lexical items. \textit{punctuation} represents the habit of punctuation use. 

We use the general SVM model from scikit-learn package~\citep{scikit-learn}.
We optimize its parameters including the $C$ and $\gamma$ and the kernel functions by training with different combinations of them and selecting the one with the best performance on the development set. 
We also normalize our feature matrices before passing it to the model.

\begin{table}[t]
  \centering
  \begin{tabular}{lccccr}
    \toprule
    \multicolumn{1}{c}{\multirow{2}{*}{\textbf{Linguistic Features}}} & \multicolumn{4}{c}{\textbf{Training Set}} & \multicolumn{1}{c}{\multirow{2}{*}{\textbf{Feature Number}}} \\
    \cmidrule(lr){2-5}
    & \textbf{All} & \textbf{50\%} & \textbf{25\%} & \textbf{10\%} & \\
    \midrule
    CFGRs (frequency $>$ 10) & 91.71 & 90.29 & 90.14 & 87 & 939 \\
    CFGRs (frequency $>$ 20) & 78.71 & 78 & 78.14 & 76.71 & 131 \\
    \midrule
    Function Words & \textbf{95.14} & \textbf{94.14} & \textbf{93.86} & \textbf{92.29} & 467 \\
    \midrule
    Top 10 Frequent Words & 75.14 & 76.29 & 75.71 & 75.43 & 10 \\
    Top 50 Frequent Words & 89.00 & 87.14 & 87.00 & 86.00 & 50 \\
    \midrule
    POS Unigrams & 90.71 & 88.86 & 88.71 & 87.71 & 45 \\
    \midrule
    Punctuation & 80 & 80.14 & 78.86 & 79.14 & 14 \\
    \midrule
    Word Unigrams & 90.71 & 87.86 & 87.57 & 86.14 & 2,409 \\
    \bottomrule
  \end{tabular}
  \caption{Document-level performance of SVM models trained on different linguistic features and experimental settings. The rightmost column shows the number of features.}
  \label{tab:ling-feat2}
\end{table}

\paragraph{Results}
As shown in Table \ref{tab:ling-feat2}, the classifier trained with function words attains an accuracy of 94.33\% with the full training set. The detector trained with POS unigrams also perform well in this task, reaching an accuracy of 87\%. The detectors using context free grammar rules and word unigrams as training feature also report an over 85\% accuracy but with more features than POS unigrams. 
Increasing the number of frequent words from 10 to 50 enhances the accuracy by nearly 10\%. 
It is interesting to note that using only 14 punctuations as features can give us an accuracy of 80\%. 

\paragraph{Analysis}
We observe that training with syntactic and stylistic features results in high performance in our detector,
which is shown in the accuracy in detector trained with POS unigrams and context free grammar rules.
In particular, the detectors trained with CFGRs and POS unigrams outperform that trained with word unigrams.
This indicates that relying on syntactic information alone, the detector can tell the differences between AIGC and human-written essays.

The detector trained with function words
achieve the highest accuracy (95.14\%, suggesting that humans and machines have different usage patterns for these words.

The performance of the model trained with the distribution of punctuation is around 80\% with only 14 features, which suggests that humans and machines use punctuations very differently. 

Overall, our results show that the SVM detector can distinguish the AIGC from human-written essays with high accuracy based on syntactic features alone. When trained with functions words as features, the SVM can achieve 95\% accuracy at the document level. 

\subsubsection{Fine-tuning RoBERTa-large for classification}

\paragraph{Methods}
To further investigate whether AI generated essays are statistically distinguishable from human essays, we fine-tune a RoBERTa-large\footnote{We use the huggingface implementation 
\citep{Wolf2019TransformersSN} from \url{https://huggingface.co/roberta-large}.}~\citep{2019RoBERTa}. 
\textcolor{black}{
Similar to how we train SVM detectors, we also train RoBERTa detectors on training sets of different granularities or different sizes to analyze the difficulties of AIGC detection.
}
We train the model for 2 epochs using the largest batch size that can fit into a single GPU with 24 GB RAM. The full set of training hyperparameters are presented in Table \ref{tab:roberta-hyperparam}. We evaluate the detector on the test set, but also report its performance on the portions of essays generated by different models, as presented in Table \ref{tab:roberta-result}.

\begin{table}
    \centering
    \begin{tabular}{ccccccc}\toprule
    $\downarrow$ Train data & gpt2-xl & babbage & curie & davinci-003 & turbo & all \\\midrule
gpt2-xl & 97.46  & 100.00  & 98.33  & 98.82  & 97.67  & 98.05 \\
babbage-001 & 98.31  & 100.00  & 98.33  & 98.82  & 98.84  & 99.19 \\
curie-001 & 97.74  & 100.00  & 99.44  & 99.41  & 99.81  & 99.33 \\
davinci-001 & 98.02  & 100.00  & 100.00  & 99.41  & 99.23  & 99.24 \\
davinci-002 & 98.31  & 99.72  & 99.45  & 99.80  & 99.42  & 99.33 \\
davinci-003 & 86.44  & 99.45  & 99.17  & 99.61  & 99.61  & 97.19 \\
turbo & 81.36  & 97.50  & 99.44  & 99.22  & 99.23  & 96.00 \\
10\% & - & - & - & - & - & 99.67 \\
25\% & - & - & - & - & - & 99.14 \\
50\% & - & - & - & - & - & 99.76 \\
all & 99.15  & 99.72  & 99.45  & 99.41  & 100.00  & 99.38 \\
    \bottomrule
    \end{tabular}
    \caption{Main results of our RoBERTa AIGC detector for document-level classification, evaluated on each test subset (each column) by accuracy. }
    \label{tab:roberta-result}
\end{table}

\paragraph{Results}
From Table \ref{tab:roberta-result}, we observe that RoBERTa easily achieves 99 accuracy in detecting AI generated essays, even when trained only on less then 10\% of the data. When directly transferred to detecting paragraph-level and sentence-level data (see Table~\ref{tab:detector:allrestest}), the model's performance drops by 23 points and 44 points respectively, but this is due to the distinct length gaps between training data and test data rather then the inherent difficulty in discriminated machine-generated sentences from human-written ones, as models trained on paragraph-level and sentence-level data scores 97.88 and 99.38 accuracy respectively on each one's i.i.d test data.

Table \ref{tab:roberta-result} also confirms that the essays generated by models from GPT-3.5 family share a similar distribution, while the essays generated by GPT-2 are likely from a different distribution. 
But we hypothesize that this is not because essays generated by GPT-2 are harder to distinguish from human essays, but rather because its essays are not as well-posed as those generated by other models, and thus introduces larger noise into the detector's training process.

\subsubsection{Zero/few-shot learning experiment of gpt-3.5-turbo as AIGC detector}
\paragraph{Methods}

We test the capability of gpt-3.5-turbo (ChatGPT) on the AIGC detection task. 
We experiment with zero/one/two-shot learning by putting zero/one/two pairs of positive and negative examples in the prompt (for details see Appendix \ref{sec:zero/few-shot}).

All evaluation is done on our validation set. Since the performance is very poor, we do not further evaluate it on the test set.

\paragraph{Results}
The accuracy for gpt-3.5-turbo on the AIGC detection task is presented in Table \ref{tab:cal:gpt3.5}. The results show that gpt-3.5-turbo performs poorly on the AIGC detection task. Under the zero-shot scenario, the model classifies almost all essays as AI-generated. Therefore, the accuracy of zero-shot is close to 50\%. 
Under the one-shot/two-shot scenarios, the average accuracy for the six pairs of prompts is also roughly 50\%, suggesting that perhaps this task is still too difficult for the model in a few-shot setting. 
The model also has poor performance at paragraph-level classification, and we do not further experiment on sentence-level evaluation. 

\begin{table}[H]
    \centering
    \begin{tabular}{ccc}
        \toprule
        \textbf{} & \textbf{Doc} & \textbf{Para} \\\midrule %
        Zero-shot & 50.33 & 43.28 \\
        One-shot & 44.56 & 36.47 \\
        Two-shot & 51.66 & 37.81 \\
        \bottomrule
        \end{tabular}
    \caption{Accuracy for zero/few-shot scenario of document and paragraph level on dev set.}
    \label{tab:cal:gpt3.5}
\end{table}

\subsubsection{Out-of-distribution performance}

To investigate the generalization ability of AIGC detectors, four aforementioned detectors are evaluated on the OOD dataset on the  doc-, para-\footnote{We don't paragraph human essays for CLEC does not provide explicit notations for paragraphs.}, and sent-level, including 
(1) our RoBERTa trained on the training set of ArguGPT dataset, 
(2) the SVM model trained on the features of \textit{function words}, 
(3) GPTZero (version 2023-06-12), and 
(4) RoBERTa released by \citeauthor{guo2023close}. 
The results are presented in Table~\ref{tab:ood:results}. 

We first make three general observations:

\begin{itemize}
\setlength\itemsep{.1em}
    \item It is much easier for detectors to detect OOD human essays---all detectors have 90+\% accuracy---than machine essays---some detectors have poor performance, at around 50\%.
    \item GPTZero has the best performance on the human OOD test set with document-level accuracy at 100.00\% and sentence level at 96.92\%.
    \item Our RoBERTa trained on the ArguGPT dataset has the best performance on the machine OOD test set (doc: 97.00\%; para: 93.13\%; sent: 83.57\%).
\end{itemize}

\paragraph{Results on the human OOD test set}

We observe that the performance on the human OOD test set is much better than the performance on the machine OOD test set.
Apart from the RoBERTa fine-tuned on other text genres~\citep{guo2023close} which has a sent-level accuracy of 60.60\%, all other four detectors have the accuracy of 94\%+ for human OOD test set at both sent- and doc-level, among which GPTZero is the best detector for human OOD essays (doc: 100.00\%; sent: 96.92\%).

\begin{table}[t]
    \centering
    \begin{subtable}[t]{\linewidth}
    
    \resizebox{\textwidth}{!}{
    \begin{tabular}{ccccccccc}
        \\
        \multicolumn{9}{c}{\textbf{Machine}} \\\toprule
        \multirow{2}{*}{Model} & \multirow{2}{*}{Level} & \multicolumn{5}{c}{Sub-corpus} & Overall & \multirow{2}{*}{ID acc./$\Delta\downarrow$} \\\cmidrule(lr){3-7}
         & & turbo & gpt-4 & claude & bloomz & flan-t5 &  OOD acc. & \\\midrule
        \multirow{3}{*}{RoBERTa} & doc & \textbf{99.67} & \textbf{100.00} & \textbf{97.00} & \textbf{95.67} & \textbf{92.67} & \textbf{97.00} & 99.71/2.71 \\
         & para & \textbf{98.85} & \textbf{95.82} & \textbf{90.33} & \textbf{79.27} & 75.67 & \textbf{93.13} & 98.71/5.58 \\
         & sent & \textbf{97.01} & \textbf{92.83} & \textbf{83.81} & \textbf{63.85} & 77.80 & \textbf{83.57} & 97.26/13.69 \\\hline
        \multirow{3}{*}{Best SVM} & doc & 85.00 & 88.00 & 75.00 & 60.00 & 53.00 & 72.20 & 94.00/21.80 \\
         & para & 83.80 & 60.69 & 59.61 & 39.00 & 46.00 & 64.43 & 89.42/24.99 \\
         & sent & 72.65 & 57.83 & 56.14 & 16.00 & 28.00 & 53.13 & 78.33/25.20 \\\hline
        \multirow{3}{*}{GPTZero} & doc & 94.00 & 32.00 & 11.00 & 54.00 & 76.00 & 53.40 & 95.42/42.02 \\
         & para & 94.27 & 50.00 & 21.16 & 56.09 & 84.00 & 57.72 & 94.06/36.34 \\
         & sent & 96.77 & 56.25 & 22.52 & 62.13 & \textbf{87.50} & 65.37 & 96.57/31.20 \\\hline
        \multirow{3}{*}{\citet{guo2023close}} & doc & 80.00 & 15.00 & 30.00 & 84.00 & 87.00 & 59.20 & 94.00/34.80 \\
         & para & 90.83 & 49.86 & 47.25 & 76.21 & \textbf{87.00} & 64.67 & 92.42/27.75 \\
         & sent & 88.08 & 59.86 & 59.92 & 61.87 & 79.88 & 69.87 & 87.19/17.32 \\\bottomrule
         \end{tabular}
         }
         \caption{Accuracy on the machine OOD test set. \texttt{turbo}: gpt-3.5-turbo; \texttt{claude}: claude-instant; \texttt{bloomz}: bloomz-7b; \texttt{flan-t5}: flan-t5-11b.}
         \end{subtable}

    \begin{subtable}[t]{\linewidth}
    
    \resizebox{\textwidth}{!}{
    \begin{tabular}{ccccccccc}
        \multicolumn{9}{c}{\textbf{Human}} \\\toprule
        \multirow{2}{*}{Model} & \multirow{2}{*}{Level} & \multicolumn{5}{c}{Sub-corpus} & Overall & \multirow{2}{*}{ID acc./$\Delta\downarrow$} \\\cmidrule(lr){3-7}
         & & st2 & st3 & st4 & st5 & st6 & OOD acc. & \\\midrule
        \multirow{3}{*}{RoBERTa} & doc & 95.33 & 99.67 & \textbf{100.00} & 97.33 & \textbf{100.00} & 98.47 & 99.05/0.58 \\
         & para & - & - & - & - & - & - & - \\
         & sent & 94.64 & \textbf{95.65} & 96.64 & 94.75 & 89.22 & 93.20 & 90.93/-2.27 \\\hline
        \multirow{3}{*}{Best SVM} & doc & 92.00 & 91.00 & 95.00 & 97.00 & 99.00 & 94.80 & 96.29/1.49 \\
         & para & - & - & - & - & - & - & - \\
         & sent & 92.89 & 90.01 & 92.00 & 89.75 & 81.61 & 87.91 & 83.25/-4.66 \\\hline
        \multirow{3}{*}{GPTZero} & doc & \textbf{100.00} & \textbf{100.00} & \textbf{100.00} & \textbf{100.00} & \textbf{100.00} & \textbf{100.00} & 98.28/-1.72 \\
         & para & - & - & - & - & - & - & - \\
         & sent & \textbf{98.09} & 94.17 & \textbf{99.75} & \textbf{96.00} & \textbf{95.61} & \textbf{96.92} & 96.57/-0.35 \\\hline
        \multirow{3}{*}{\citet{guo2023close}} & doc & 96.00 & \textbf{100.00} & 99.00 & \textbf{100.00} & \textbf{100.00} & 99.00 & 85.71/-13.29 \\
         & para & - & - & - & - & - & - & - \\
         & sent & 71.11 & 62.64 & 71.46 & 64.82 & 46.98 & 60.60 & 58.23/-2.37 \\\bottomrule
         \end{tabular}
    }
    \caption{Accuracy on the human OOD test set. \texttt{st2}: senior high school; \texttt{st3}: juniors in Univ.; \texttt{st4}: seniors in Univ.; \texttt{st5}: juniors of English major; \texttt{st6}: seniors of English major.}
    \end{subtable}
    
    \caption{Accuracy on out-of-distribution (OOD) test set. Overall OOD acc.: overall OOD accuracy. ID acc.: in-domian accuracy. $\Delta$: difference between OOD and ID accuracy. Note that versions of GPTZero used in OOD and ID are different for we can only access to the API of the latest version. }
    \label{tab:ood:results}
\end{table}

\paragraph{Results on machine OOD test set}

When we turn to results on OOD essays written by generative models, we see that the RoBERTa finetuned on ArguGPT training data achieves exceptionally high accuracy at all three levels (doc: 97\%, para: 93\%, sent: 83\%).
At the document level, the performance is only 2 percentage points below the in-distribution performance, while at the sentence level, we see a drop of roughly 13\% from in-distribution performance (see last column of Table~\ref{tab:ood:results}). 

Our best model of SVMs performs second best at the document-level. 
Nevertheless, the SVM model has the lowest accuracy in prediction at the sentence level. 

On the other hand, the two detectors that are not specifically trained for argumentative essay detection---GPTZero and RoBERTa of \citeauthor{guo2023close}---exhibit surprisingly  poor performance, with an accuracy of 53.40\% and 59.20\% respectively. For the RoBERTa from \citet{guo2023close}, this could be attributed to the fact that their training data do not contain argumentative essays, and that all the texts in their training data are generated by one model---gpt-3.5-turbo.

The performance of GPTZero is particularly unsatisfactory for essays generated by gpt-4 (32\% at doc-level) and claude-instant (11\%). Thus one should be cautious when using GPTZero for detection essays written by AI models other than gpt-3.5-turbo. 
We also note that GPTZero has better performance on finer-grained prediction (i.e. sent-level $>$ para-level $>$ doc-level). 
However, there is a reverse trend for the ArguGPT-finetuned  RoBERTa, namely it performs best at the document-level, but worst at sent-level. 

For essays generated by different models, detectors have drastically different performance. 
For essays generated by gpt-3.5-turbo, all detectors have a similar performance compared to the in-distribution (ID)  evaluation on the ArguGPT test set, except for the SVM and RoBERTa from \citet{guo2023close} which has a roughly 10\% gap between the OOD and ID performance. 
However, for essays generated by gpt-4 and claude-instant, it becomes extremely difficult for two off-the-shelf detectors, GPTZero and \citet{guo2023close}, with their accuracy between   11\% and 32\%. On the other hand, our RoBERTa finetuned on the ArguGPT training data shows almost no performance drop from ID to OOD evaluation at the document level (acc: 95+\%). 
For essays generated by the two smaller language models bloomz-7b and flan-t5-11b, we observe that the two off-the-shelf detectors have a better performance, with accuracy between 54\% and 87\% at the doc-level.

\paragraph{Discussion and implications} 

Our experiments on the OOD test set have several implications. 

First, when evaluating AIGC detectors, it is necessary to construct a more comprehensive evaluation set that covers text generated by multiple models, as detection accuracy varies dramatically for text generated by different models.
\textcolor{black}{In our experiments, 
detectors have much better performance of predicting gpt-3.5-turbo than other models.
However, as these generative models quickly update and new models emerge, an evaluation set should ideally cover as many models as possible so as to reflect the actual detection performance. 
}

Second, transferring to detect AIGC generated by a different model might be more difficult than transferring to a different text genre. This is manifest in the ID and OOD performance of the RoBERTa by \citet{guo2023close}, which is fine-tuned on text of 5 genres generated solely by gpt-3.5-turbo: while it has 80+\% accuracy on detecting argumentative essays written by gpt-3.5-turbo, the performance drops to 15\% and 30\% respectively for essays generated by gpt-4 and claude-instant.
Our results suggest that text produced by different models may have distinctive textual features that can be challenging for transfer learning. This resonates with our first point where a more comprehensive evaluation set is necessary.

Third, it is easier for the detectors to identify human essays than machine essays. Our OOD results show that the detectors have higher performance on human essays, suggesting that the human essays in our OOD set are more or less homogeneous to the ID data, whereas the machine essays are likely from a genuine different distribution.

\section{Related work}\label{sec:related}
\subsection{The evolution of large language models}
Since \citet{2017Transformer} proposed Transformer, a machine translation model that relies on self-attention, language models have kept advancing at an unprecedented pace in recent years. The field has seen innovative ideas ranging from the pretraining-finetuning paradigm~\citep{2018GPT,2018BERT} and larger-scale mixed-task training~\citep{2019T5} to implicit~\citep{radford2019language-gpt2} and explicit~\citep{2021FLAN,2021T0}  multitask learning and in-context learning~\citep{brown2020language-gpt3}. Some works scaled language models to hundreds of billions of parameters~\citep{2021Gopher,2022MT-NLG,2022PaLM}, while others reevaluated scaling laws~\citep{2020scaling,2022Chinchilla} and trained smaller models on larger amount of higher-quality data~\citep{2023PaLM2,2023Phi-1.5}. Some trained them to follow natural~\citep{2021NaturalInstructions}, supernatural~\citep{2022supernatural}, and unnatural~\citep{2022unnatural} instructions. Others investigated their abilities in reasoning with chain-of-thought~\citep{Wei2022CoT,Wang2022CoT,Kojima2022Cot}.

More recently, OpenAI's GPT models have become front-runners in language models that displayed exceeding performances in various tasks. GPT-2 is a decoder-only auto-regressive language model with 1.5B parameters presented by \citeauthor{radford2019language-gpt2}, who pretrained the model on 40GB of Web-based text and found it to demonstrate preliminary zero-shot generation abilities on downstream tasks such as task-specific language modeling, commonsense reasoning, summarization, and translation. GPT-3 (i.e. davinci, \citealp{brown2020language-gpt3}) is an enlarged version of GPT-2 with 175B parameters and pre-trained on a larger corpus that mainly consists of CommonCrawl.
InstructGPT~\citep{ouyang2022training}, also known as text-davinci-001, is a more advanced version of GPT-3 finetuned by both instructions and reinforcement learning with human feedback (RLHF). Text-davinci-002 applied the same procedures to Codex (i.e. code-davinci-002, \citealp{2021Codex}), a variant of GPT-3 that is further pretrained on code using the same negative log-likelihood objective as language modeling. Text-davinci-003 improved text-davinci-002 further by introducing Proximal Policy Optimization (PPO). GPT-3.5-turbo, more commonly known as ChatGPT, is a variant of text-davinci-003 optimized for dialogues. Besides all these large language models, OpenAI has also released a series of smaller models, including curie, babbage, and ada, each one smaller in size and faster at inference.

\subsection{Human evaluation of AIGC}

In Natural Langauge Generation (NLG) tasks, human evaluation is often considered to be the gold standard, for the goal of NLG is to generate readable texts for human~\citep{celikyilmaz2020evaluation}. Human subjects are often asked to perform a Turing test~\citep{machinery1950computing} in order to evaluate the human-likeness of machine-generated texts.

\citet{brown2020language-gpt3} asked 80 participants to identify news articles generated by language models with different parameter sizes from human-written articles. 
The results show that the accuracy of human identifying model generated articles dropped to a chance level with large language models like GPT-3. 
\citet{clark_all_2021} recruited 780 participants to identify texts generated by the GPT-2 and the davinci models. The results indicate that the accuracy of human participants is around 50\%, though the accuracy becomes higher after training. 
They suggested that human could misunderstand and underestimate the ability of machine. 
\citet{uchendu2021turingbench} also reported that human subjects were able to identify machine-generated texts only at a chance level.

\citet{dou_is_2022} proposed a framework that used crowd annotation to scrutinize model-generated texts. The results show that there are some gaps between human-authored and machine-generated texts; for instance, human-authored texts contain more grammatical errors while machine-generated texts are more redundant. \citet{guo2023close} asked participants to compare the responses from both human experts and ChatGPT in various domains; their results suggest that people find answers generated by ChatGPT generally more helpful than those generated by human experts in certain areas. They also identified some distinctive features of ChatGPT that are different from human experts, including that ChatGPT is more likely to focus on the topic, display objective and formal expression with less emotion.  

In summary, it is difficult for human subjects to distinguish texts generated by advanced large language models from human-written texts.

\subsection{AIGC detector}

AIGC detection is a text classification task that aims to distinguish machine-generated texts from human-written texts. Even before the release of ChatGPT, researchers had presented many works on machine-generated text detection. \citet{gehrmann2019gltr} developed GLTR, a tool that applies statistical methods to detect machine-generated texts and improve human readers' accuracy in identifying machine-generated text detection. \citet{zellers2019defending} developed Grover, a large language model that was trained to detect machine-generated news articles. \citet{uchendu-etal-2020-authorship} trained several simple neural models that aim to not only distinguish machine-generated texts from human-written texts, i.e. the Turning Test (TT), but furthermore, identify which NLG method generated the texts in question, i.e. the Authorship Attribution (AA) tasks. 
They found that those neural models (especially fine-tuned RoBERTa) performed reasonably well on the classification tasks, though texts generated by GPT2, FAIR, and GROVER were more difficult than other models. \citet{uchendu2021turingbench} introduced a TuringBench dataset for TT and AA tasks. Results from their preliminary experiments indicate that better solutions, including more complex models, are needed in order to meet the challenges in text classification.

The recent surge of AIGC, as a result of advances in large language models, further motivates researchers to explore AIGC detectors. GPTZero is one of the earliest published ChatGPT-generated content detectors. OpenAI also published its own detector\footnote{\url{https://openai.com/blog/new-ai-classifier-for-indicating-ai-written-text}} after the prevalence of ChatGPT. \citet{guo2023close} trained detectors basing on deep learning and linguistic features to detect ChatGPT-generated texts in Q\&A domain. \citet{mitrovic2023chatgpt} trained perplexity-based and DistilBert fine-tuned detectors. \citet{mitchell2023detectgpt} developed a zero-shot detector. 

\subsection{AIGC in the education domain}

Large language models, especially OpenAI's GPT-3 and ChatGPT, have displayed impressive performance in academic and professional exams. \citet{Choi2023law-exam} used ChatGPT to produce answers on Law School's exams consisting of 95 multiple choice questions and 12 essay questions. The results show that ChatGPT achieved a low but passing grade in all those exams. \citet{zhang2022ml-exam} evaluated the answers generated by large language models, including Meta's OPT, OpenAI's GPT-3, ChatGPT, and Codex, on questions from final exams of undergraduate-level Machine Learning courses. The results show that large language models achieved passing grades on all selected final exams except one and scored as high as a B+ grade. They also used those models to generate new questions for Machine Learning courses and compared the quality, appropriateness, and difficulty of those questions with human-written questions via an online survey. The results suggest that AI-generated questions, though slightly easier than human-written questions, were comparable to human-written questions regarding quality and appropriateness. \citet{Kung2023-med-exam} analyzed ChatGPT's performance on the United States Medical Licensing Exam (USMLE), a set of three standardized tests that assess expert-level knowledge and are required for medical licensure in the US. The findings indicate that ChatGPT performed at or near the passing threshold of 60\% and was able to provide comprehensible reasoning and valid clinical insights. 

Additionally, the latest GPT-4 model has shown remarkable progress, further surpassing the previous GPT models. Its performance is comparable to that of humans on most professional and academic tests, and it even achieved a top 10\% score on a simulated bar exam~\citep{openai2023gpt4}. One exception to GPT models' stellar performances across various subjects in exams is their ability to solve mathematical problems. \citet{frieder2023math-exam} evaluated ChatGPT's performance on a data set consisting of mathematical exercises, problems, and proofs. The results show that ChatGPT failed to deliver high-quality proofs or calculations consistently, especially for those in advanced mathematics.

Furthermore, the remarkable performances of large language models suggest that they have the potential to influence various aspects of the education domain. \citet{Kung2023-med-exam} pointed out that large language models such as ChatGPT could facilitate human learners in a medical education setting. \citet{Zhai2022-writing} examined ChatGPT's ability in academic writing and concluded that ChatGPT could deliver a high-quality academic paper quickly with minimal input. \citet{article} presented the potential benefits of ChatGPT for research, including increased efficiency for researchers and the possibility of reviewing articles. \citet{thorp2023} argued that ChatGPT could provide high-accuracy English translation and proofreading to non-native English speakers, narrowing the disadvantages they face in the field of science. 

However, the increasing use and broad implications of large language models have raised ethical concerns regarding the accuracy and authenticity of generated content, academic integrity, and plagiarism. \textit{Science} has updated its policies to prohibit ChatGPT from being listed as an author, and AI-generated texts should not be included in published work~\citep{thorp2023}. \citet{gao2022fake-abstract} used ChatGPT to generate research abstracts based on existing abstracts from medical journals. They found that plagiarism detectors failed to identify AI-generated abstracts. Though the AIGC detector can distinguish generated and human-written, original abstracts, human reviewers found it challenging to identify generated abstracts. \citet{Alkaissi2023-fake-writing} used ChatGPT to write about the pathogenesis of medical conditions. The results show that while ChatGPT could search for related information and produce credible and coherent scientific writings, the references it provided contained unrelated or even fabricated publications. Both teams of authors recommend that editorial processes for journals and conferences should integrate AIGC detectors and clearly disclose the use of these technologies.

\section{Conclusion}
\label{sec:conclusion}

To sum up, in this paper we have
(1) compiled a human-machine balanced corpus of argumentative essays, 
(2) hired 43 English instructors and conducted human evaluation and identification of machine-authored essays, 
(3) analyzed and compared human and machine essays from the perspectives of syntactical and lexical complexity, 
and (4) built and evaluated classifiers for detection of GPT-generated essays.

Results of human evaluation suggest that English instructors have difficulty in distinguishing machine-generated essays from human-written ones, with an accuracy of 0.6163 in round 1, which can be improved to 0.6767 when they are shown the correct answers for round 1 and asked to reflect on the characteristics of machine-written essays.
They have a higher accuracy of identifying low-level human essays and high-level machine essays, suggesting that 
participants anticipate that GPT-generated essays should have higher quality than human-written ones.

In our linguistic analyses, we find that GPT models produce sentences with more complex syntax than human, but the diction and vocabulary of human essays tend to be more diverse. 
What should be noted is that more advanced language models generally rank higher on synatctic and lexical complexity. 

Finally, it is easy for machine-learning classifiers to capture the differences between human-authored and machine-generated essays, as an in-domain RoBERTa-large model achieves 99+\% accuracy on the document-level  and 93+\% on sentence-level classification. 
Results of SVM classifiers show that using surface (POS unigram) and deep (context-free grammar) syntactic features alone yields high performance, suggesting that human and machine essays can be distinguished by these structural characteristics. 
\textcolor{black}{However, it is difficult for most detectors (except our document-level RoBERTa) to transfer to out-of-distribution generative models.}

We hope our study can be helpful to ESOL instructors in identifying AIGC, and also useful for NLP practitioners interested in human evaluation and textual analysis of AIGC. 

\section*{Acknowledgments}
We thank Rui Wang, Yifan Zhu, and Huilin Chen 
for discussions on early drafts of the paper and their help in the human evaluation. We are also grateful to all participants in our human evaluation experiment. 
This project is supported by 
the Humanities and Social Sciences Grant from the Chinese Ministry of Education (No. 22YJC740020) awarded to Hai Hu.

\begin{multicols}{2}
\bibliography{main}
\end{multicols}

\appendix

\section*{Appendix}

\section{Examples of deleted gpt2-generated texts}
\label{ap:gpt2:deleted}

Here we present some essays generated by gpt-2 that were manually removed.

\paragraph{Q\&A}
Q: What is the most important thing you have learned in your life so far that you would like to pass on to the next generation? // A: I have learnt that there is no such thing as a free lunch...

\paragraph{Providing URL}
If you're not sure what to write about, you can always ask a friend or family member to help you out. You can also use a free online writing service like \url{WriterDuet.com} to get started. 

\paragraph{Nonsense}
Refuse to answer \% \% ALL 18-34 35-44 45-54 55-64 65-74 75-84 85-94 95-100 \% of all voters Clinton (D) Trump (R) Johnson (L) Stein (G) Undecided (V) Refused (N) Total All voters 45.4\% 35.0\% 27.3\% 25.6\% 21.5\% 17.1\% 13.9\% 10.8\% 8.2\%...

\begin{CJK*}{UTF8}{gbsn}
\paragraph{Speaking Chinese}
What do you think is the most important thing you can do to improve the quality of life for yourself and your loved ones? 龚属思想让我们没有办法自己的情况下,但这个世界的经验也不会觉得到过去的...
\end{CJK*}

\section{Automatic scoring details}

\subsection{Scores given by YouDao automated scoring systems under different criteria}
\label{ap:auto:crit}

\begin{table}[H]
    \centering
    \resizebox{\linewidth}{!}{
    \begin{tabular}{c|cccccc}
    \toprule
    \textbf{Essay label} & \textbf{babbage-001} & \textbf{curie-001} & \textbf{text-davinci-001} & \textbf{text-davinci-002} & \textbf{text-davinci-003} & \textbf{gpt-3.5-turbo} \\ \midrule
    01 & 96.8 & 106.5 & 89.4 & 93.1 & 105.6 & 100.1 \\
    02 & 89.1 & 83.1 & 87.4 & 92.0 & 106.5 & 106.5 \\
    03 & 106.5 & 106.2 & 83.4 & 106.5 & 102.0 & 106.5 \\
    04 & 78.5 & 98.8 & 81.2 & 85.2 & 103.9 & 106.4 \\
    05 & 78.7 & 93.5 & 104.9 & 101.1 & 106.5 & 105.6 \\
    06 & 100.3 & 101.2 & 89.8 & 105.2 & 106.5 & 103.3 \\ \bottomrule
    \end{tabular}}
    \caption{Essay scores under the criterion of CET6.}
    \label{tab:auto:score:cet6}
\end{table}

\begin{table}[H]
    \centering
    \resizebox{\linewidth}{!}{
    \begin{tabular}{c|cccccc}
    \toprule
    \textbf{No.} & \textbf{babbage-001} & \textbf{curie-001} & \textbf{text-davinci-001} & \textbf{text-davinci-002} & \textbf{text-davinci-003} & \textbf{gpt-3.5-turbo} \\ \midrule
    01 & 15 & 16 & 15 & 16 & 20 & 19 \\
    02 & 15 & 14 & 15 & 16 & 16 & 20 \\
    03 & 19 & 19 & 13 & 19 & 16 & 20 \\
    04 & 12 & 17 & 11 & 15 & 19 & 20 \\
    05 & 14 & 16 & 19 & 18 & 19 & 19 \\
    06 & 18 & 16 & 15 & 18 & 19 & 20 \\ \bottomrule
    \end{tabular}}
    \caption{Essay scores under the criterion of Graduate.}
    \label{tab:auto:score:graduate}
\end{table}

\begin{table}[H]
    \centering
    \resizebox{\linewidth}{!}{
    \begin{tabular}{c|cccccc}
    \toprule
    \textbf{No.} & \textbf{babbage-001} & \textbf{curie-001} & \textbf{text-davinci-001} & \textbf{text-davinci-002} & \textbf{text-davinci-003} & \textbf{gpt-3.5-turbo} \\\midrule
    01 & 99 & 96 & 86 & 97 & 98 & 99 \\
    02 & 89 & 85 & 85 & 97 & 99 & 94 \\
    03 & 98 & 97 & 90 & 97 & 99 & 99 \\
    04 & 90 & 98 & 88 & 95 & 91 & 98 \\
    05 & 89 & 99 & 97 & 96 & 98 & 97 \\
    06 & 94 & 93 & 88 & 98 & 91 & 98 \\\bottomrule
    \end{tabular}}
    \caption{Essay scores under the criterion of Default.}
    \label{tab:auto:score:default}
\end{table}

\section{Details for added prompts}
\label{ap:prompts}
\begin{table}[H]
    \centering
    \resizebox{\linewidth}{!}{
    \begin{tabular}{c|p{12cm}}
    \toprule
    \textbf{Prompt label} & \multicolumn{1}{c}{\textbf{Detail}} \\ \midrule
    01 & Do you agree or disagree? Use specific reasons and examples to support your answer. Write an essay of roughly 300/400/500 words. \\ \hline
    02 & Do you agree or disagree? It is a test for English writing. Please write an essay of roughly 300/400/500 words. \\ \hline
    03 & Do you agree or disagree? Pretend you are the best student in a writing class. Write an essay of roughly 300/400/500 words, with a large vocabulary and a wide range of sentence structures to impress your professor. \\ \hline
    04 & Do you agree or disagree? Pretend you are a professional American writer. Write an essay of roughly 300/400/500 words, with the potential of winning a Nobel prize in literature. \\ \hline
    05 & Do you agree or disagree? From an undergraduate student's perspective, write an essay of roughly 300/400/500 words to illustrate your idea. \\ \bottomrule
    \end{tabular}}
    \caption{Details for 5 added prompts in our pilot study}
    \label{tab:prompt:detail}
\end{table}

\section{Instructions for human experiment} \label{app:instructions:human:exp}

\centerline{English Essays Judgment Experiment}

Welcome to the English Judgment Experiment by Shanghai Jiao Tong University!\footnote{The instruction is translated by DeepL and proofread by one of the authors.} The task of this experiment is to determine whether an essay is written by a human or a machine (AI model). You will complete the experiment in two rounds with the same experimental setup. In each round, 10 English essays will be judged, with some written by humans and others by machines. You will see the essay question and the writing from TOEFL (Test of English as a Foreign Language). The essay questions are randomly the same or different.

A variety of AI models, regardless of their performances, were used in this experiment, including ChatGPT(i.e., get-3.5-turbo). The human essays are written by TOEFL test takers from around the world. They have different native languages and varying levels of English language proficiency. The purpose of this experiment is to understand the sensitivity of English teachers to machine-generated essays and to improve their ability to recognize the machine-generated essays in the second experiment through the first round of tests.

To determine the authorship of an essay, you should choose one of the following six options:

1 = Definitely written by humans, 

2 = Probably written by humans, 

3 = Likely written by humans, 

4 = Likely written by machines, 

5 = Probably written by machines,

6 = Definitely written by machines.

You can give reasons for your judgment (optional).

When you have finished your answer, click  ``submit." You will see the correct answers on a new page, and you should summarize the characteristics of machine-generated essays (required).
After completing the first round, you will move on to the second round, which will follow the same procedure as the first round. We would like you to apply the experience gained from the first round to improve your accuracy in the second round.

Your goal is to make as many correct judgments as possible, and each correct answer will increase the payout by 1 RMB. Please do not take screenshots during the experiment! All results will be disclosed after they are compiled. Please use a computer for the experiment.

Estimated time: 20-30 minutes.

Experimental reward: 40 RMB (basic reward) with bonus (number of correct answers*2 RMB). As there are 20 questions in total, the maximum reward is 80 RMB. Please fill in your Alipay account and the actual name of the account holder on the last page (required for funding reimbursement). You can also choose not to be paid if you want to keep your Alipay information private. The payment will be made within one month after the completion of the experiment.

Notice: 
\begin{itemize}
\item To ensure that the experiment runs properly, please do not share your username or password with others, and do not use multiple devices to log into your account at the same time during the experiment. All non-up-to-date sessions will be forced to log out. 
\item The purpose of the user name is to number the experiment data. The password is to prevent others from attacking this site and has no effect on the experiment. You can set your password at will. 
\item At the end of the experiment, please fill in the information related to your teaching experience. Your information and answers will be kept confidential and used only for the analysis of experimental results. 
\item If you have any questions or suggestions, please contact argugpt@163.com (preferred) or hu.hai@sjtu.edu.cn.
\end{itemize}

\centerline{Questionnaire for Human Experiment}
Thank you for taking the English Essays Judgment Experiment by Shanghai Jiao Tong University!

Please fill out the following personal information. Your information will be kept confidential and used only for the analysis of experimental results.

\textbf{You are currently:}
\begin{itemize}
    \item Undergraduate student majoring in English (including translation and linguistics)
\item Master's student in English
    \item PhD student in English
    \item Assistant professor or lecturer in English (including public and foreign affairs, translation, English specialization, etc.)
    \item Professor of English
    \item High school English teacher
    \item Middle school English teacher
    \item English teacher of training institution
    \item Other: (Please fill in the blank)
\end{itemize}

\textbf{Which of the following statement best describes your situation:}
\begin{itemize}
    \item I am an English major, but have basically never corrected student essays
    \item I am an English major and have corrected student essays (as a teaching assistant or by marking essays for CET-4  or CET-6)
    \item I am a teacher and have been teaching English courses for 1-5 years
    \item I am a teacher and have been teaching English courses for 5-10 years
    \item I am a teacher and have been teaching English courses for 10-20 years
    \item I am a teacher and have been teaching English courses for 20-30 years
    \item I am a teacher and have been teaching English courses for 30 years or more
    \item Other: (Please fill in the blank)
\end{itemize}

\textbf{Do you know about AI writing tools such as ChatGPT or XieZuoCat?}
\begin{itemize}
    \item No, I haven't heard any of them.
    \item Yes, but I don't know how AI writing tools work.
    \item Yes, and I know how they work.
\end{itemize}

\textbf{Have you ever used AI writing tools such as ChatGPT or XieZuoCat before?}
\begin{itemize}
\item No, never.
\item Yes, occasionally.
\item Yes, usually.
\item If you have used AI tools, please describe the scenarios and purposes of your use.
\end{itemize}

\textbf{How do you usually mark your compositions? (multiple choice)}
\begin{itemize}
    \item Corrected by myself
    \item Corrected by an teaching assistant
    \item PiGaiWang
    \item Youdao
    \item Grammarly
    \item iwrite
\end{itemize}

\textbf{Would you consider using AI tools in your teaching?}
\begin{itemize}
\item Very reluctant
\item Reluctant
\item Not necessarily
\item Willing
\item Very willing
\end{itemize}

\textbf{Do you think your ability to identify machine-generated essays is improved through this experiment?}
\begin{itemize}
\item Yes
\item No idea
\item No
\end{itemize}

\textbf{Do you think AI writing poses a challenge to foreign language teaching?}
\begin{itemize}
\item Please fill in the blank
\end{itemize}

\section{Training details of RoBERTa}\label{sec:training-details}
The same set of hyperparameters presented in Table \ref{tab:roberta-hyperparam} are used to train document, paragraph, and sentence level classifiers. These hyperparameters are empirically selected, without any tuning on the validation set.
\begin{table}[H]
    \centering
    \begin{tabular}{lc}
    \toprule
        \textbf{Hyperparam} &  \\\midrule
        Learning Rate & 7.5e-6 \\
        Batch Size & 5 \\ 
        Weight Decay & 0.01 \\ 
        Epochs & 2 \\ 
        LR Decay & linear \\ 
        Warmup Ratio & 0.05 \\
    \bottomrule
    \end{tabular}
    \caption{Hyperparameters for fine-tuning RoBERTa to build AIGC detector.}
    \label{tab:roberta-hyperparam}
\end{table}

\section{Zero/few-shot prompts for gpt-turbo-3.5 in AIGC detection experiment}
\label{sec:zero/few-shot}

In order to reduce the impact of different prompt essays on the performance of the model, we select a total of six pairs of essays as our example essays in one/two-shot scenarios. In each pair, two essays are generated/written under the same prompt. For human essays, we select two essays from each level (low, medium, and high). As their counterpart in machine essays, we select six essays generated by different GPT models and divided them into three groups corresponding to the three different levels. With the six pairs of example essays, we have six sets of results for one-shot and three for two-shot. 
Details of the prompts are show in Table~\ref{tab:pmpt:gpt3.5}.

\begin{table}[H]
    \centering
    \resizebox{\linewidth}{!}{
    \begin{tabular}{c|p{17.7cm}}
        \toprule
        \# Shot & Prompts \\\midrule
        zero-shot & \makecell[l]{Question: Is the following content written by human or machine? Please reply human or machine. \\ Essay: \textless{}test\_essay\textgreater \\ Answer: }\\
        \hline
        one-shot & \makecell[l]{Question: Is the following content written by human or machine? Please reply human or machine. \\ Essay: \textless{}human\_essay\textgreater \\ Answer: Human \\ Question: Is the following content written by human or machine? Please reply human or machine. \\ Essay: \textless{}machine\_essay\textgreater \\ Answer: Machine \\ Question: Is the following content written by human or machine? Please reply human or machine. \\ Essay: \textless{}test\_essay\textgreater \\ Answer:} \\
        \hline
        two-shot & \makecell[l]{Question: Is the following content written by human or machine? Please reply human or machine. \\ Essay: \textless{}human\_essay\_1\textgreater \\ Answer: Human \\ Question: Is the following content written by human or machine? Please reply human or machine. \\ Essay: \textless{}machine\_essay\_1\textgreater \\ Answer: Machine \\ Question: Is the following content written by human or machine? Please reply human or machine. \\ Essay: \textless{}human\_essay\_2\textgreater \\ Answer: Human \\ Question: Is the following content written by human or machine? Please reply human or machine. \\ Essay: \textless{}machine\_essay\_2\textgreater \\ Answer: Machine \\ Question: Is the following content written by human or machine? Please reply human or machine. \\ Essay: \textless{}test\_essay\textgreater \\ Answer:}  \\
        \bottomrule
    \end{tabular}
    }
    \caption{Prompt format for zero/few-shot scenario}
    \label{tab:pmpt:gpt3.5}
\end{table}

\section{Linguistic analysis}\label{sec:app-ling-analysis}

\subsection{Lexical richness metrics}
\begin{table}[H]
\centering
\resizebox{\linewidth}{!}{
\begin{tabular}{llll}\toprule
\multicolumn{1}{l}{Dimension} & \multicolumn{1}{l}{Measure} & \multicolumn{1}{l}{Code} & \multicolumn{1}{l}{Formula} \\\midrule
Lexical Density & Lexical Density & LD & $N_{lex}/N$\\\midrule
\multirow{5}{*}{Lexical Sophistication} & Lexical Sophistication-I & LS1 & $N_{slex}/N_{lex}$ \\
 & Lexical Sophistication-II & LS2 & $N_{s}/T$\\
 & Verb Sophistication-I & VS1 & $T_{sverb}/N_{verb}$ \\
 & Verb Sophistication-II & VS2 & $T^{_{2}}_{sverb}/N_{verb}$ \\
 & Corrected VS1 & CVS1 & $T_{sverb}/\sqrt[]{2N_{verb}}$\\\midrule
\multirow{17}{*}{Lexical Variation} & Number of Different Words & NDW & $T$ \\
 & Ndw (First 50 Words) & NDW-50 & $T$ in the first 50 words of sample \\
 & Ndw (Expected Random 50) & NDW-ER50 & Mean $T$ of 10 random 50-word samples \\
 & Ndw (Expected Sequence 50) & NDW-ES50 & Mean $T$ of 10 random 50-word sequences \\
 & Type-Token Ratio & TTR & $T/N$ \\
 & Mean Segmental TTR (50) & MSTTR-50 & Mean TTR of all 50-word segments \\
 & Corrected TTR & CTTR & $T/\sqrt{2N}$ \\
 & Root TTR & RTTR & $T/\sqrt{N}$ \\
 & Bilogarithmic TTR & LogTTR & $LogT/LogN$ \\
 & Uber Index & Uber & $Log^{2}N/Log(N/T)$ \\
 & Lexical Word Variation & LV & $T_{lex}/N_{lex}$ \\
 & Verb Variation-I & VV1 & $T_{verb}/N_{verb}$ \\
 & Squared VV1 & SVV1 & $T^{2}_{verb}/N_{verb}$ \\
 & Corrected VV1 & CVV1 & $T_{verb}/\sqrt{2N_{verb}}$ \\
 & Verb Variation-Ii & VV2 & $T_{verb}/N_{lex}$ \\
 & Noun Variation & NV & $T_{noun}/N_{lex}$ \\
 & Adjective Variation & AdjV & $T_{adj}/N_{lex}$\\
 & Adverb Variation & AdvV & $T_{adv}/N_{lex}$\\
 & Modifier Variation & ModV & $(T_{adj}+T_{adv})/N_{lex}$\\\bottomrule
\end{tabular}
}
\caption{Lexical measures, replicated from Table 1 \& 2 of \citet{Lu2012TheRO}. }\label{tab:lexical:measures}
\end{table}

Results for lexical density and lexical sophistication are shown in Table~\ref{tab:lexical:density:sophistication}. 
Results for lexical variation can be found in Table~\ref{tab:lexical:variation}.

\subsection{Lexical richness of ArguGPT Corpus}

\begin{table}[H]
\centering
\begin{tabular}{lcccccc}
\toprule
 & Lexical Density & \multicolumn{5}{c}{Lexical Sophistication} \\\hline
 & Lexical Density & \multicolumn{2}{c}{Lexical Sophistication} & \multicolumn{3}{c}{Verb Sophistication} \\\hline
 & ld & ls1 & ls2 & vs1 & vs2 & cvs1 \\\hline
human-low & 0.51 & 0.18 & 0.17 & 0.11 & 0.50 & 0.40 \\
human-medium & 0.50 & 0.17 & 0.17 & 0.11 & 0.61 & 0.46 \\
human-high & 0.51 & 0.21 & 0.22 & 0.14 & 1.51 & 0.75 \\
human-average & 0.50 & 0.19 & 0.19 & 0.12 & 0.87 & 0.53 \\\midrule
gpt2-xl & 0.46 & 0.12 & 0.11 & 0.06 & 0.11 & 0.16 \\
text-babbage-001 & 0.49 & 0.12 & 0.12 & 0.05 & 0.09 & 0.15 \\
text-curie-001 & 0.49 & 0.15 & 0.15 & 0.07 & 0.19 & 0.25 \\
text-davinci-001 & 0.50 & 0.16 & 0.15 & 0.07 & 0.22 & 0.27 \\
text-davinci-002 & 0.50 & 0.15 & 0.15 & 0.08 & 0.29 & 0.32 \\
text-davinci-003 & 0.52 & 0.21 & 0.21 & 0.10 & 0.57 & 0.48 \\
gpt-3.5-turbo & 0.53 & 0.23 & 0.24 & 0.13 & 0.97 & 0.64\\\bottomrule
\end{tabular}
\caption{Lexical density and lexical sophistication of ArguGPT Corpus.}\label{tab:lexical:density:sophistication}
\end{table}

\begin{table}[H]
\centering
\resizebox{\linewidth}{!}{
\begin{tabular}{lccccccccccccccccccc}
\toprule
 & \multicolumn{19}{c}{Lexical Variation} \\\hline
 & \multicolumn{4}{c}{Number of Different Words} & \multicolumn{6}{c}{Type-Token Ratio} & \multicolumn{9}{c}{Type-Token Ratio of Word Class} \\\hline
 & ndw & ndwz & ndwerz & ndwesz & ttr & msttr & cttr & rttr & logttr & uber & lv & vv1 & svv1 & cvv1 & vv2 & nv & adjv & advv & modv \\\hline
human-low & 96.53 & 36.24 & 36.73 & 36.48 & 0.49 & 0.73 & 4.74 & 6.70 & 0.86 & 16.94 & 0.62 & 0.68 & 11.86 & 2.37 & 0.17 & 0.60 & 0.11 & 0.08 & 0.20 \\
human-medium & 131.03 & 37.28 & 37.97 & 37.72 & 0.45 & 0.75 & 5.38 & 7.61 & 0.86 & 17.69 & 0.62 & 0.68 & 16.85 & 2.85 & 0.17 & 0.59 & 0.12 & 0.08 & 0.20 \\
human-high & 190.76 & 37.79 & 38.98 & 38.65 & 0.44 & 0.77 & 6.36 & 8.99 & 0.86 & 19.31 & 0.62 & 0.70 & 25.79 & 3.52 & 0.17 & 0.59 & 0.12 & 0.08 & 0.20 \\
human-average & 139.44 & 37.10 & 37.89 & 37.62 & 0.46 & 0.75 & 5.49 & 7.77 & 0.86 & 17.98 & 0.62 & 0.68 & 18.17 & 2.92 & 0.17 & 0.59 & 0.12 & 0.08 & 0.20 \\\midrule
gpt2-xl & 81.14 & 36.52 & 37.18 & 36.23 & 0.56 & 0.73 & 4.59 & 6.49 & 0.88 & 18.57 & 0.74 & 0.76 & 9.87 & 2.16 & 0.20 & 0.72 & 0.15 & 0.09 & 0.23 \\
text-babbage-001 & 72.20 & 34.60 & 33.32 & 33.62 & 0.40 & 0.68 & 3.75 & 5.30 & 0.82 & 12.94 & 0.51 & 0.55 & 6.56 & 1.76 & 0.13 & 0.48 & 0.11 & 0.06 & 0.18 \\
text-curie-001 & 88.49 & 36.34 & 34.94 & 35.59 & 0.41 & 0.72 & 4.19 & 5.93 & 0.83 & 14.12 & 0.54 & 0.59 & 8.89 & 2.07 & 0.14 & 0.50 & 0.11 & 0.07 & 0.18 \\
text-davinci-001 & 99.36 & 36.51 & 35.42 & 36.40 & 0.40 & 0.73 & 4.43 & 6.27 & 0.83 & 14.58 & 0.54 & 0.58 & 9.99 & 2.19 & 0.14 & 0.51 & 0.11 & 0.06 & 0.18 \\
text-davinci-002 & 99.59 & 36.26 & 35.62 & 36.35 & 0.41 & 0.73 & 4.45 & 6.30 & 0.83 & 14.72 & 0.54 & 0.58 & 10.32 & 2.23 & 0.14 & 0.51 & 0.11 & 0.07 & 0.18 \\
text-davinci-003 & 137.61 & 36.68 & 36.69 & 37.78 & 0.37 & 0.75 & 5.00 & 7.07 & 0.83 & 15.24 & 0.51 & 0.55 & 14.34 & 2.63 & 0.13 & 0.49 & 0.11 & 0.05 & 0.16 \\
gpt-3.5-turbo & 166.42 & 38.68 & 38.07 & 38.88 & 0.39 & 0.78 & 5.65 & 7.99 & 0.84 & 16.89 & 0.54 & 0.59 & 18.81 & 3.03 & 0.14 & 0.52 & 0.11 & 0.05 & 0.16\\\bottomrule
\end{tabular}
}
\caption{Lexical variation of ArguGPT Corpus.}\label{tab:lexical:variation}
\end{table}

\end{document}